\documentclass[sn-apa]{sn-jnl}
\usepackage[utf8]{inputenc}
\usepackage{dirtytalk}
\usepackage{amsmath}
\usepackage{amssymb}
\usepackage{amsfonts}
\usepackage{graphicx}
\usepackage{csquotes}
\usepackage{textcomp}
\usepackage{booktabs}
\usepackage{lipsum}
\usepackage{adjustbox}
\usepackage{xcolor}
\usepackage{array}
\usepackage{exscale}
\usepackage{color}
\usepackage{soul}
\usepackage{relsize}
\usepackage{float}
\usepackage{rotating}
\usepackage{subcaption}
\usepackage{bbold}
\usepackage{multirow}
\usepackage{scalerel}
\usepackage{algorithm}
\usepackage{algpseudocode}
\usepackage{graphics}
\usepackage{epsfig}
\usepackage{xfrac}
\usepackage{etoolbox}
\makeatletter

\def\BState{\State\hskip-\ALG@thistlm}
\makeatother
\DeclareMathOperator*{\argmax}{arg\,max}

\newcolumntype{L}[1]{>{\raggedright\let\newline\\\arraybackslash\hspace{0pt}}m{#1}}
\newcolumntype{C}[1]{>{\centering\let\newline\\\arraybackslash\hspace{0pt}}m{#1}}
\newcolumntype{R}[1]{>{\raggedleft\let\newline\\\arraybackslash\hspace{0pt}}m{#1}}

\definecolor{googleblue}{rgb}{0.259,0.522,0.957}
\definecolor{googlegreen}{rgb}{0.060,0.620,0.350}
\definecolor{googlered}{rgb}{0.859,0.267,0.220}
\definecolor{bleudefrance}{rgb}{0.19,0.55,0.91}

\jyear{2023}%

\theoremstyle{thmstyleone}

\theoremstyle{thmstyletwo}

\theoremstyle{thmstylethree}

\begin{document}

\title[Contextually Aware Intelligent Control Agents for Heterogeneous Swarms]{Contextually Aware Intelligent Control Agents for Heterogeneous Swarms}

\author*[1]{\fnm{Adam J.} \sur{Hepworth}}\email{a.j.hepworth@unsw.edu.au}

\author[1]{\fnm{Aya S.M.} \sur{Hussein}}\email{a.hussein@adfa.edu.au}

\author[1,2]{\fnm{Darryn J.} \sur{Reid}}\email{darryn.reid@dst.defence.gov.au }

\author[1]{\fnm{Hussein A.} \sur{Abbass}}\email{h.abbass@unsw.edu.au}

\affil*[1]{\orgdiv{School of Engineering and Information Technology}, \orgname{University of New South Wales}, \orgaddress{\street{Northcott Drive}, \city{Canberra}, \state{ACT}, \postcode{2600}, \country{Australia}}}

\affil[2]{\orgname{Defence Science and Technology Group}, \orgaddress{\city{Adelaide}, \state{SA}, \postcode{5111}, \country{Australia}}}

\abstract{An emerging challenge in swarm shepherding research is to design effective and efficient artificial intelligence algorithms that maintain a low-computational ceiling while increasing the swarm\textquoteright s abilities to operate in diverse contexts. We propose a methodology to design a context-aware swarm-control intelligent agent. The intelligent control agent (shepherd) first uses swarm metrics to recognise the type of swarm it interacts with to then select a suitable parameterisation from its behavioural library for that particular swarm type. The design principle of our methodology is to increase the situation awareness (i.e. information contents) of the control agent without sacrificing the low-computational cost necessary for efficient swarm control. We demonstrate successful shepherding in both homogeneous and heterogeneous swarms.}


\maketitle

\section{Introduction}

Contemporary approaches to swarm guidance and control often assume that swarm agents are homogeneous in their response to external influence vectors. This manifests in the design of control algorithms, such as herding, often operating directly on the raw positional data of swarm agents to compute influence vectors. Herding-based models, such as shepherding, have been implemented for over 25 years, with classic control methods typically operating on simple transformations of raw data~\cite{Hasan2022:FlockNav}. Swarm shepherding is an example of a swarm control herding-based method where one or more external actuators (sheepdogs) operate on low-level information by calculating primitive statistical features from raw data. These models often use static behaviour selection policies for the control agent to guide a swarm to a goal location~\cite{9659082}. As a biologically-inspired approach to swarm control, shepherding has applications across different domains, such as the guidance and control of crowds~\cite{10.1007/978-3-642-34381-0_48}, herding biological animals~\cite{Paranjape2018}, guiding teams of uncrewed system (UxS)~\cite{Hepworth2021:ARS}, and controlling a group of robotic platforms~\cite{s17122729, Cowling:2010:AHS:3014666.3014668}.

While the design principle of swarm systems aims at lowering the computational cost for transforming sensor data into actionable information (S2AI) and executing the logic for the actionable-information-to-actuation (AI2A) cycle, they can only operate successfully in simple environments and are known to be fragile when these assumptions are violated~\cite{9256255}. Common examples of when they fail to include the presence of noise in sensed data and increased diversity/heterogeneity in the swarm contexts they need to operate within.

There are typically nested sets of assumptions common between most swarm control models, such as employing binary sensor models, linear decision-action boundaries, event-based decision models, homogeneous swarm agents, and predefined swarm parameters~\cite{Long2020:Comprehensive}. Furthermore, the most prevalent approaches to shepherding-based swarm control employ rule-based behaviour policies, with relatively few examples of swarm control systems that adopt behaviour-switching policies~\cite{Hussein:AAMAS22}, for example~\cite{9533722, 9387150, 9659082}. While the simplicity of reactive models offers many benefits in their application, more complex individual and team behaviours may be within reach if we enhance specific aspects of an agent\textquoteright s cognition~\cite{doi:10.1098/rstb.2020.0309}.

Few studies consider requirements such as heterogeneous swarm agents, particularly swarms where behaviours of the collective and individuals may change over time~\cite{Hepworth2020:Footprints}, or employ context-aware approaches~\cite{Abbass2018Testing}. Instead, the predominant research direction follows work such as~\cite{Strombom:2014}, considering homogeneous settings with single control agents. Swarms constituted by agents who exhibit heterogeneous behaviours are known to increase the complexity and unpredictability of swarm control~\cite{Ozdemir2017:shepherding}. For approaches that consider disparate agent behaviours and characteristics with distinct influence impacts on a swarm, the control agent requires an ability to discriminate these states and subsequently determine the most appropriate behaviours to use. The recognition of these contexts is an open research problem.

Recognising heterogeneous agent contexts and swarm situations is vital in settings where one or more individual swarm members may disproportionately impact the collective behaviour~\cite{JOLLES2020278}, limiting the effectiveness of a swarm control agent. A control agent needs to identify the agent effectively and swarm dynamics, select an appropriate response/tactic and act on this tactic. The primary research question for this work asks \emph{how can context be used to adapt behaviour?} Specifically, we want to know \emph{how to adaptively select herding tactics based on recognisable swarm characteristics.} We aim to understand how a swarm control agent could use context to select an effective control strategy to better influence the swarm\textquoteright s behaviour. Our objective is to integrate context awareness into the decision-making process for an adaptive reactive rule-based swarm control agent, increasing the number of swarm situations the control agent can act on. This paper's primary contribution is presenting a context-aware intelligent swarm control agent.

Rule-based shepherding is often criticised for its inability to address dynamic and unknown swarm contexts; we address this research gap by providing a context-aware system to augment rule-based shepherding control agent. The agent is enabled with the two primary abilities of context recognition and tactic selection. These capabilities are delivered in two phases: an offline strategy evaluation and an algorithm for online strategy selection. Our methodology is presented using these two phases. We first evaluate the performance of various herding strategies across a range of homogeneous and heterogeneous swarms in a shepherding context. We seek to answer the question \emph{how do different herding strategies impact mission performance across different shepherding contexts?} In the second phase, we consider the context-aware adaptation of the control agent\textquoteright s behaviour by monitoring the swarm in real-time to optimally select a herding strategy based on the inferred swarm characteristics. Put simply, we want to know \emph{how can context be used to influence behaviour selection?} Therefore, in addition to the algorithmic contributions presented in this paper, the two conceptual contributions of this work are:
\begin{enumerate}
    \item By systematically assessing diverse shepherding behaviours in both homogeneous and heterogeneous swarm scenarios, we demonstrate a methodology to form a library of situations and behaviours to support the AI2A cycle. We show that the appropriate parameterisations of the same atomic actions used for homogeneous swarms leads to effective and efficient control of non-homogeneous swarms with disparate heterogeneity settings.
    \item By designing new metrics for swarm performance together with information markers from our previous work, we propose new models to support the S2AI cycle. This contribution culminates in a multi-agent architecture for context awareness to augment a swarm control agent in shepherding settings, validating our model concerning an established reactive rules-based control agent model. Our architecture integrates high-level information to consider the situation of a swarm, enabling context assessment and behaviour selection at different temporal resolutions. Furthermore, we demonstrate that a modest adaptation of existing rules-based decision structures and behaviour models successfully herds homogeneous and non-homogeneous swarm settings.
\end{enumerate}

In Section~\ref{sec:background}, we review approaches to swarm control and influence, surveying essential model-type formulations and critically assessing the underlying assumptions of these models. Then, in Section~\ref{sec:AgentDesign}, we present our context-aware system and formulate the problem space. Next, sections~\ref {sec:S2AI} and~\ref{sec:AI2A} cover the systematic analysis of the problem space to derive the S2AI information, followed by the AI2A design that transforms that analysis into context-aware actions. Finally, we conclude the paper in Section~\ref{sec:Conclusion} with a discussion on future research opportunities for context-aware swarm control.

\section{Methods of Swarm Control}\label{sec:background}

Swarm control encompasses a wide range of approaches to swarming, typically categorised as centralised or decentralised, with examples of each in both simulations and on physical systems~\cite{Long2020:Comprehensive}. The most prominent algorithms to swarm control in the literature are based on the notions of influence zones utilising reactive decision models, typically characterised with spatial features~\cite{PhysRevLett.75.1226}. In addition, biologically-inspired approaches feature prominently within the literature, with researchers recreating interactions, for instance, the natural flocking of birds~\cite{Reynolds1987Boids}, the predator-prey interactions of sheep and sheepdogs~\cite{Strombom:2014}, and hunting methods of a wolf pack~\cite{Hu2022:Wolfpack}.

Current approaches to swarm control often focus on problems where swarms of agents possess homogeneous sensors, reactive decision processes, and limited action sets~\cite{Hepworth2020:Footprints}. Centre-of-mass, also known as centre-of-gravity, based approaches are a standard method to control a swarm, with few assumptions around agent homogeneity or influence distribution within the swarm. Current approaches are often data-driven, operating on spatially-derived features to calculate the next move of an agent, for instance~\cite{9504706, 9256255}. Biologically-inspired models use derivations from empirical research as the basis for agent parameterisations.

One approach to centralised control is shepherding, \enquote{inspired by sheepdogs and sheep, where the shepherding problem can be defined as the guidance of a swarm of agents from an initial location to a target location}~(pg.523)~\cite{Long2020:Comprehensive}. Proposals to solve the shepherding challenge include bio-inspired algorithms, heuristic-based rule algorithms, and machine learning solutions, including neural networks and reinforcement learning approaches~\cite{Hasan2022:FlockNav}. The predominant approach to extensions of rule- or heuristic-based methods employ arc, line or circle formations, usually relying on the centre of mass and exact position of agents to conduct guidance. Few studies research the impact of limited sensing ranges and the use of only local information by the control agent, for instance~\cite{9504706, 9256255}.

\subsection{Models of Shepherding}

Shepherding behaviours can be defined as \enquote{those in which external agents control the movement of a swarm of agents}~(pg.2)~\cite{Hussein:AAMAS22}. \cite{Strombom:2014} introduce a shepherding model of interactions between a flock of sheep and a sheepdog based on empirical field trial data. The heuristic model employs a self-propelled particle approach to reproduce the attraction and repulsion interactions of the two agent types. The model describes $N$ swarm agents (sheep, $\pi$) placed within an $L \times L$ area (paddock) with $M$ control agents (sheepdogs, $\beta$). Swarm agent behaviour is generated as a force vector that combines attraction to their local centre of mass, repulsion from other $\pi$ and repulsion from $\beta$ agents. The task of $\beta$ agents is to move $\pi$ agents to a particular location, a goal area. Two behaviours are presented for the control, being \emph{collect} and \emph{drive}, derived from side-to-side movements introduced by~\cite{1308924}. The control agent collects members of the swarm within a radius ($f(N)$) and then drives the swarm toward the goal location. The mission is considered complete when the swarm centre of mass is within a distance $\delta$ from the goal, signifying that the flock is within the target goal area.

There are different implementations of shepherding algorithms for swarm control, with examples in the literature over the past two decades. These usually operate on the raw positional data of swarm agents, or simple transformations of these, relying on diverse control models for shepherding. For example, \cite{Cowling:2010:AHS:3014666.3014668} employ a \enquote{hierarchical and stack-based finite state machine to receive and act on commands by setting action states and calculating appropriate paths}~(pg.4). \cite{8228268} discuss a hybrid control method based on positional data that combines formation and collection control methods. The method proposes V-formation control to \enquote{trace a V-shaped notch toward the target position at all times}~(pg.2427), where Collecting via Centre (CvC) sees the shepherding agent collect with one of three position-cases, being left, right or centre. \cite{TsunodaY.2018Aols}~\enquote{proposed a movement law that approaches the sheep farthest from the goal in the flock and shows its superiority to the movement laws of Vaughan et al. and Strombom et al. by simulation.}~\cite{Himo2022:HeterogeneousResponse}.

\cite{9782555} propose a control agent employing a distributed collecting algorithm that does not require the centre of mass to be calculated and that drives using a density-based method instead of using the convex hull of the flock. The approach has robustness over classic methods, which may be able to address adversarial agents, such as non-cooperative members or threats external to the swarm. \cite{Auletta2022:Herding} propose a set of local control rules for a small group of herding agents to collect and herd a swarm. The primary difference in the model proposed considers the situation where some swarm agents do not possess the ability to cooperate with other members of the swarm (unable to flock), increasing the complexity of the problem space for the swarm control agent. The solution developed is implemented as a distributed approach where each herder agent selects a strategy based on local feedback, driving their decision selection for what targets to follow.

\cite{Varadharajan22} consider a variation of the shepherding problem in which a particular pattern is maintained during movement to the goal, noted as crucial for applications such as nanomedicine or smart materials or where spatial configurations may have functional implications. Also considered is that the shepherds can modulate interaction forces between the sheep. \cite{Himo2022:HeterogeneousResponse} consider the response where some swarm agents are unresponsive to the herding agent. These heterogeneous traits of some swarm members are parameterised through modification of an agent\textquoteright s repulsion forces to the control agent. An unresponsive agent is encoded with a lower response weight than the standard swarm agent and a non-responsive agent repulsion force coefficient is set to zero. The solution proposed is to collect all responsive agents around an unresponsive agent and then guide the group to the goal area. Note that this assumes that the unresponsive agent is responsive to the agents around it.

Existing rule-based algorithms lack adaptability to respond in changing environments (obstacles) or with changing swarms (non-homogeneous); however, they are shared, in part, due to their simplicity and potential application in real robotic systems~\cite{Zhang2022:Herding}. Recent shepherding models consider using learning-based control algorithms, addressing some shortfalls of rule-based algorithms. For example, \cite{9387150} proposes a method of shepherding to herd agents amongst obstacles using a deep reinforcement learning approach. \cite{Hussein:AAMAS22} consider curriculum-based reinforcement learning and propose an algorithm that demonstrates superior performance to that of the classic rules-based agent across all training and task types. \cite{9504706} introduce a graph-based approach that promotes cohesion between swarm members, improving shepherding performance and mission success outcomes.

\subsection{Context-Awareness}

Context can be defined as any information that can be used to characterise the situation of an entity relevant to the task at hand. In shepherding, context could be defined by characteristics of the control agent, swarm agents, environment, and user priorities~\cite{Onto4MAT:2022}. Context-sensitive systems \enquote{are those that respond to changes in their environment}~(pg.306)~\cite{10.5555/647985.743843} by utilising context awareness to optimise performance in non-stationary environments. Context awareness in swarming has recently observed an increase in research interest, ranging from context integrated to modulate particular behaviour attributes to sharing context for strategy cooperation between agents. \cite{Singh2019:Modulation} propose a force function to optimise energy use by the shepherding agent. Context information modulates the force vector in a classic shepherding formulation following the rule structure of \cite{Strombom:2014} for the multi-shepherd situation.

\cite{Mohanty2020:ContextSwarm} investigate heterogeneous swarms under typical communication constraints, assuming no direct communication between swarm agents. Proposed is a context-aware Deep Q-Network framework to \enquote{obtain communication-free cooperation between the robots in the swarm}~(pg.1), where the objective is to complete a mission as quickly as possible. \cite{9533722} note that few approaches address how the herding agent learns, with most reinforcement learning models unsuccessful in their application. An imitation learning approach is presented that learns from experts. \cite{9663179} presents an architecture to manage elements of context awareness across distinct domains on multi-agent systems. Six common context requirements are outlined: distribution and node heterogeneity; timing requirements; adaptability; scalability; security and privacy; and service availability.

\subsection{Heterogeneity in Swarms}

Heterogeneity is common in natural systems for agent morphology and behaviour~\cite{Kengyel:2015}. One type of heterogeneous behaviour is where swarm members have predetermined roles, each taking on a different function within the swarm. This contrasts with a task-switching view, where agents display heterogeneous behaviours depending on the needs of the collective. For task-switching heterogeneity, a swarm agent must be capable of switching roles on demand.

\cite{Kengyel:2015} utilised four agent parameterisations. The authors consider heterogeneous swarms where agents are given a predetermined behaviour, and task switching is not allowed, a static non-task-switching system, investigating if a heterogeneous swarm can outperform a homogeneous swarm under certain conditions.

\cite{10.1016/j.neucom.2016.06.064} propose a model for heterogeneous flocking in a distributed setting, considering the role of leaders within a swarm. As a decentralised model, the impact on a swarm control agent is not considered in this study. \cite{Goel:2019} further note that for heterogeneity manifesting through models of leadership and followership, \enquote{a precise measure of influence using leaders or predators or a combination of leaders and predators to achieve the mission is not adequately studied}~(pg.1).

In a centralised swarm control setting, \cite{Jang2018:LocalInfoSwarm} propose a guidance method for heterogeneous swarms that allocate agents to bins, developing an empirical agent distribution to guide the swarm. \cite{8267142} propose a multi-target tracking control method for heterogeneous swarms that also employs a distributed approach to generate stable group behaviours.

\subsection{Critical Assessment of Swarm Control Methods}

Common amongst methods identified throughout the survey of \cite{Long2020:Comprehensive} is that contemporary methods for swarm guidance employ similar control model formulations. These include that the control agent decision model is often rule-based, initiating primitive force-vector behaviours based on raw data or simple transformation of this. Long et al. highlight that \enquote{finding the right set of weights in a context is a dependent problem}~(pg.534), further going on to discuss that modulation of these weights may be required in different environments structures or in settings where uncertainty exists in either the sensing or decision-action output of an agent. For settings where the swarm agents have increased intelligence, \enquote{activity recognition becomes a vital problem that needs to be resolved}~(pg.534), in which \enquote{little research has taken place on this topic in the animal world}~(pg.534). A discussion on heterogeneous swarms is absent, however, with \cite{Long2020:Comprehensive} indicating that future shepherding control systems could be complemented by expanding on current methods with additional capabilities, such as goal planning and path planning, as well as considering new behavioural sets to enhance dynamics complexity.

Selecting the correct behaviour under increased uncertainty can lead to \enquote{difficulty in determining which behaviours are required to achieve effective shepherding and when they should be used}~(pg.2634)~\cite{9659082}. For example, where flocking agents require interaction rules to generate flocking behaviours, control agents require steering rules to generate herding behaviours. Predominantly, proposed control methods assume that the control agent has access to perfect information for the position of the swarm agents~\cite{s17122729}. In this paper, we require adaptive behaviours to cater for uncertainty in the swarm behaviour and the state; a summary of models and swarm control implementations is contained in Table~\ref{table:SwarmScenarios}.

\begin{table*}[h!]
    \centering
    \def\arraystretch{1.25}
    \resizebox{\textwidth}{!}{%
    \begin{tabular}{llllC{5em}C{3em}}
        \toprule
        \textbf{Study}                                                              & \textbf{Setting}    & \textbf{Decision Model}      & \textbf{Framework} & \textbf{Heterogeneous} & \textbf{Context-Aware}    \\ \midrule
        \cite{Reynolds1987Boids}                                    & Flocking                  & Rule-based                        & Decentralised              &            &            \\
        \cite{PhysRevLett.75.1226}                                  & Flocking                  & Rule-based                        & Decentralised              &            &            \\
        \cite{1308924}                                              & Shepherding               & Rule-based                        & Single agent               &            &            \\
        \cite{1570636}                                              & Shepherding               & Rule-based                        & Multi agent                &            &            \\
        \cite{Cowling:2010:AHS:3014666.3014668}                     & Shepherding               & Finite State Machine              & Single agent               &            &            \\
        \cite{10.1007/978-3-642-34381-0_48}                         & Shepherding               & Rule-based with path planning     & Single agent               &            &            \\
        \cite{Strombom:2014}                                        & Shepherding               & Rule-based                        & Single agent               &            &            \\
        \cite{Masehian2015}                                         & Shepherding               & Rule-based                        & Decentralised              &            &            \\
        \cite{Kengyel:2015}                                         & Honeybee                  & Evolutionary                      & Decentralised              & \checkmark &            \\
        \cite{10.1016/j.neucom.2016.06.064}                         & Flocking                  & Adaptive Controller               & Decentralised              & \checkmark &            \\
        \cite{s17122729}                                            & Shepherding               & Rule-based                        & Multi agent                &            &            \\
        \cite{Ozdemir2017:shepherding}                              & Shepherding               & Rule-based                        & Single agent               &            &            \\
        \cite{Jang2018:LocalInfoSwarm}                              & Fish School               & Markov chain                      & Decentralised              & \checkmark &            \\
        \cite{8716084}                                              & Shepherding               & Rule-based                        & Single agent               &            &            \\
        \cite{TsunodaY.2018Aols}                                    & Shepherding               & Rule-based                        & Single agent               & \checkmark &            \\
        \cite{Goel:2019}                                            & Shepherding               & Rule-based                        & Multi agent                &            &            \\
        \cite{8267142}                                              & Flocking                  & Rule-based                        & Decentralised              & \checkmark &            \\
        \cite{Singh2019:Modulation}                                 & Shepherding               & Rule-based                        & Single agent               &            & \checkmark \\
        \cite{9070871}                                              & Non-Bio                   & Soft-max regression               & Decentralised              &            & \checkmark \\
        \cite{10.1007/978-3-030-36708-4_54}                         & Shepherding               & Deep reinforcement learner        & Single agent               &            &            \\
        \cite{Hepworth2020:Footprints}                              & Shepherding               & Rule-based                        & Single agent               & \checkmark &            \\
        \cite{9504706}                                              & Shepherding               & Connectivity-aware rule-based     & Single agent               &            &            \\
        \cite{9256255}                                              & Shepherding               & Rule-based                        & Multi agent                &            &            \\
        \cite{Mohanty2020:ContextSwarm}                             & Non-Bio                   & Deep learning                     & Decentralised              & \checkmark & \checkmark \\
        \cite{9533722}                                              & Shepherding               & Imitation learning                & Single agent               &            & \checkmark \\
        \cite{9387150}                                              & Shepherding               & Reinforcement learner (RL)        & Single agent               &            &            \\
        \cite{9659082}                                              & Shepherding               & Neuro-evolutionary NN-RL          & Single agent               &            &            \\
        \cite{9782555}                                              & Shepherding               & Rule-based                        & Multi agent                &            &            \\
        \cite{Auletta2022:Herding}                                  & Shepherding               & Rule-based                        & Multi agent                &            &            \\
        \cite{Varadharajan22}                                       & Shepherding               & Rule-based                        & Multi agent                &            &            \\
        \cite{Himo2022:HeterogeneousResponse}                       & Shepherding               & Rule-based                        & Single agent               & \checkmark &            \\
        \cite{Hussein:AAMAS22}                                      & Shepherding               & Reinforcement learner             & Single agent               &            &            \\
        \cite{Zhang2022:Herding}                                    & Shepherding               & Rule-based                        & Multi agent                &            &            \\
        \cite{Abpeikar2022:TransferLearn}                           & Flocking                  & Rule-based                        & Distributed                &            &            \\
        \cite{Hu2022:Wolfpack}                                      & Wolf Pack                 & Rule-based                        & Multi agent                &            &            \\
        \bottomrule
    \end{tabular}
    }
    \caption{Summary of studies discussed, highlighting the consideration of model setting, decision model type, swarm heterogeneity and control agent context-awareness. Not all studies focus on swarm control; however, all employ a model of swarming within their solution. Rule-based solutions are those typically with precise mathematical formulations.}
    \label{table:SwarmControlLiterature}
\end{table*}

In uncomplicated settings, limited cognition elements are required, and the swarm control agent can often act on the raw positional information. However, in settings where the homogeneity assumption is relaxed, enhanced cognition of the swarm control agent may be required to determine agent characteristics and understand how these manifest as a source of control imbalance in the swarm. This point of differentiation in approaches to typical explorations of swarm control models allows us to consider a more comprehensive range of scenarios that could include adversarial agents, additional collaborating control agents and environmental complexities.

\cite{Zhang2022:Herding} highlights that typical herding patterns require two behaviours, collecting and driving. A central assumption is that swarm agents are collected and, once aggregated, driven toward a goal location. Focusing on the task execution sequence may introduce fragility to reactive shepherding models, limiting the possible strategies for a control agent to implement. In settings where the swarm constituent agents possess heterogeneous properties, it may be desirable to collect and drive sub-groups of agents, or one at a time, to the goal location. In such a scenario, the swarm may only be collected at the goal location as the final agent arrives.

The above review suggests an open research gap for recognising swarm and swarm agent characteristics. Without such a capability, a swarm control agent cannot understand the cumulative impacts of influences on and in a swarm and subsequently act on this information to determine the most appropriate control strategy. Furthermore, this approach requires integrating contexts to develop increased situational awareness.

\section{Context-Aware Intelligence System}\label{sec:AgentDesign}

\cite{PAJARESFERRANDO201322} states that a \enquote{system is context-aware if it can extract, interpret and use context information and adapt its functionality to the current context of use.} Our motivation for a context-aware system is to provide a control agent with the capability to select and modulate autonomous behaviours for swarm control. Often rule-based swarm control agents receive information from the environment that leads to a new autonomous behaviour being selected, for instance, where a swarm agent is moving around the decision boundary between collect and drive actions. These cases can result in a decision \emph{deadlock} in which the control agent continuously changes their implemented behaviour, leading to limited or no progress towards the mission goal. Our idea is to integrate time-based autonomous behaviour modulations to maintain performance stability. We hypothesise that the combination of time-based behaviour modulation and context awareness for shepherding-based control will increase performance and decrease observed instability across the swarm in control settings.

In our previous work~\cite{Hepworth2022:SwarmAnalytics}, an observer aimed to reveal one or more hidden state aspects of the swarm. In our current work, the observer\textquoteright s role is to support the decision-making process of the shepherding agent, providing context awareness about the swarm to determine which behaviour parameterisation is best for an observed situation. In our setting, the context-aware system observes the environment to determine the situation of the swarm, characterising the type of agents and swarm identified. Upon classifying the context, the observer agent parameterises the behaviours available to the control agent to best move the swarm to the goal location. The context-aware system constrains the autonomous behaviours available to the control agent, modulating its impact on the swarm. The context-awareness system observes the environment for the period determined by the context period, re-sampling the swarm to detect changes in the environmental situation. Our context-aware augmentation aims to parameterise behaviours for the cognitive agent.

\begin{figure}
    \centering
    \includegraphics[width=0.75\textwidth]{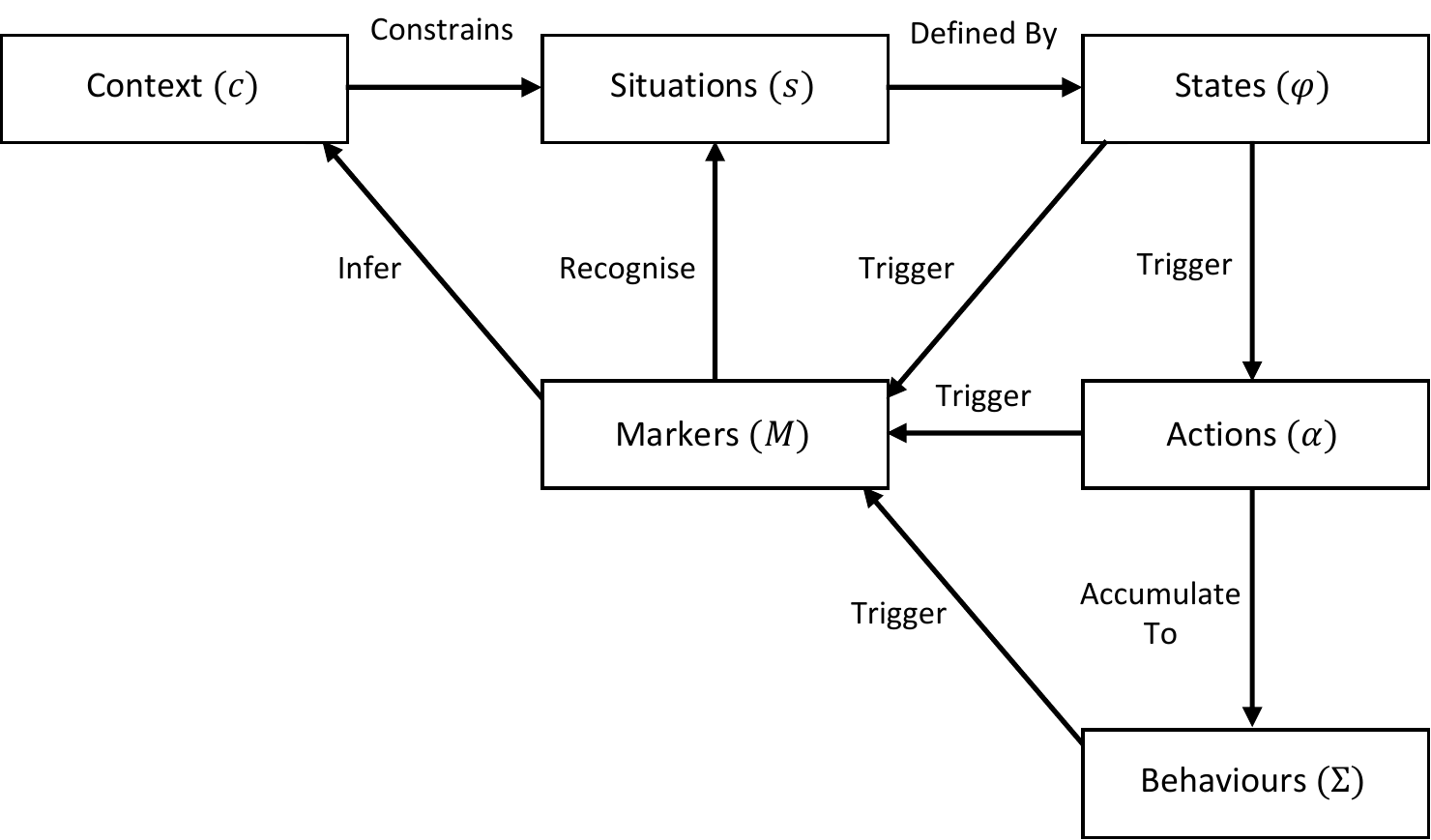}
    \caption{Conceptual linkages of definitions, expanded on that presented in~\cite {Hepworth2022:SwarmAnalytics}.}
    \label{fig:DefinitionsMapLarge}
\end{figure}

Figure~\ref{fig:DefinitionsMapLarge} conceptualises our extension of the work presented in~\cite{Hepworth2022:SwarmAnalytics}~(Figure~4) to incorporate states, actions and behaviours as aspects of the control agent system. These conceptual relationships link the system instantiated context and situations, the inference mechanism to recognise situations and contexts (information markers), and subsequent reasoning to enable an agent to act on the contexts and situations identified (states, actions and behaviours). The context constrains the situation, with states defined by the situation observed. The current state of the swarm and its agent triggers an action in the swarm control agent that accumulates into behaviours. As the markers are triggered, a context can be inferred.

\begin{figure*}[h!]
    \centering
    \includegraphics[width=\textwidth]{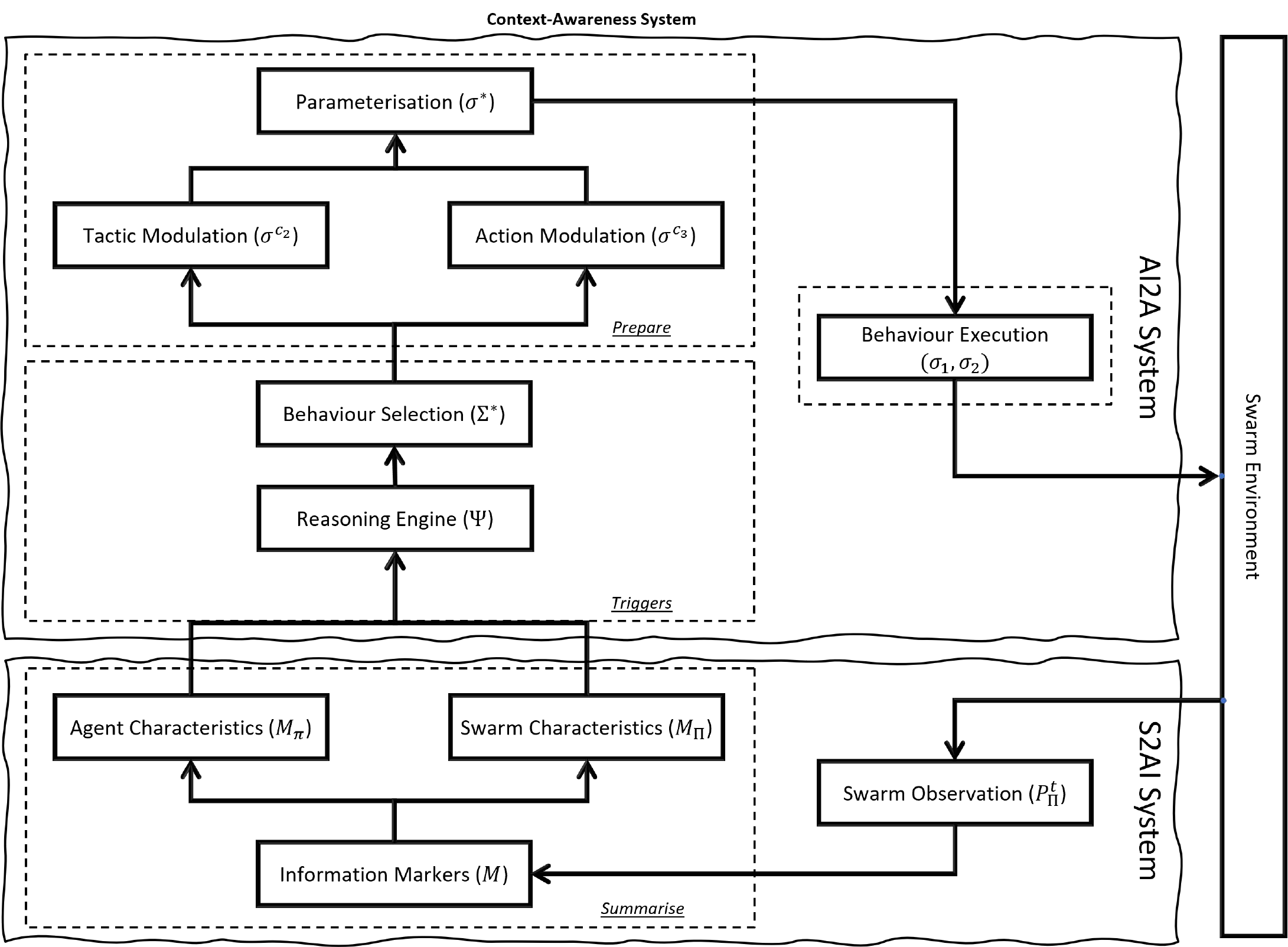}
    \caption{Conceptual architecture depicting the flow of information for the context-awareness system presented.}
    \label{fig:ControlAgentArchitecture}
\end{figure*}

We present the top-level architecture of our context-aware reasoning in Figure~\ref{fig:ControlAgentArchitecture}, depicting the flow of information for the system. The architecture consists of three primary functions, consisting of 1) summarising the swarm through information markers calculated on the positional information of agents, 2) reasoning on triggered information markers to select a particular behavioural response, and 3) preparing the agent for action by parameterising the behaviours and their properties available to the agent over the subsequent observation period.

The proposed architecture augments a classic shepherding agent by providing cognitive capabilities that operate at a level above the reactive behaviour execution, for example, as described in~\cite{Strombom:2014}. The architecture modules function as follows.

\begin{itemize}
    \item \textbf{Sensor Data to Actionable Information Phase (S2AI)}
\begin{itemize}

    \item \textbf{Summarise.} The summarise function is based on the information markers method presented in ~\cite{Hepworth2022:SwarmAnalytics}. Information markers ($\mathcal{M}$) use sensed position information of all $\pi_i$ to detect state information about $\Pi$. As described in Figure~\ref{fig:DefinitionsMapLarge}, we employ information markers in a pattern recognition function to infer contexts and recognise situations. Marker states uncover particular aspects of agent characteristics ($\mathcal{M}_{\pi_i}$), enabling type classification. At the collective level, individual states accumulate to discover swarm categories ($\mathcal{M}_\Pi$), for instance, the distribution of agents and type of homogeneity or heterogeneity observed.
\end{itemize}

    \item \textbf{Actionable Information Phase to Actuation (AI2A)}

\begin{itemize}
    \item \textbf{Trigger.} The trigger function synthesises classified individual and collective characteristics about the swarm, enabling the reasoning engine ($\Psi$) to infer the swarm contexts and situations. The reasoning engine longitudinally considers the output of information marker classifications to develop a most-likely situation scenario for the observed swarm. We use the recognised swarm situation to constrain the search space of feasible behaviours for the control agent and select the behaviours made available for execution ($\Sigma^*$).

    \item \textbf{Prepare.} The prepare function determines how often an autonomous behaviour should be executed ($\sigma^{c_2}$), as well as how frequently the selected behaviour should be re-parameterised ($\sigma^{c_3}$). The modulation of these behaviour elements enables the control agent to consider multiple periods of swarm response, designed to address the often high levels of sensorial noise previously discussed. The final aspect of the prepare function is to parameterise the control agent with the behaviours available for execution ($\sigma^*$) and the constraints for their employment ($\sigma^{c_2}$, $\sigma^{c_3}$).
\end{itemize}
\end{itemize}

Algorithm~\ref{Algo:ContextAnalytics} describes the functions discussed and depicted in Figure~\ref{fig:ControlAgentArchitecture}. The context-aware decision shepherding has two user-defined inputs, $\omega$ and $\tau$. These parameters set the observation window length ($\omega$) and proportion of overlap between each context window ($\tau \in [0,1]$), as given on Line~\ref{alg:line:1}. In our previous work, we established $\omega = 60$ and $\tau = 0.75$ for optimal $\pi_i$ and $\Pi$ recognitions \cite{Hepworth2022:SwarmAnalytics}. The context-awareness system continues to operate until the mission success criteria are achieved; the success criteria are defined as a threshold distance between the goal location and the centre of mass of the swarm, $\vert\vert P_\Pi - P_G \vert\vert\leq R_G$ (Line~\ref{alg:line:2}).

During the observation period window ($\Delta\omega$), the context-awareness system observes the coordinate position of each $\pi$, given as $P^{\Delta\omega}_{\pi_i}$ (Lines~\ref{alg:line:3}-\ref{alg:line:4}). These recorded observations are used to first calculate information markers ($\mathcal{M}$, Line~\ref{alg:line:5}) for each $\pi$, classifying each agent as a particular type ($A$, Line~\ref{alg:line:6}), as given in Table~\ref{table:AgentCharacteristics}. Finally, each probabilistic classification is summarised as a state vector ($A^*$) that reports the likelihood each agent type has is observed; the state vector $A^*$ is the empirical swarm distribution (Line~\ref{alg:line:8}).

For each known scenario in Table~\ref{table:SwarmScenarios} ($s \in S$, Line~\ref{alg:line:9}), we then calculate the $L2$-norm distance between $s$ and $A^*$, given as $\vert\vert s - A^* \vert\vert_2$ (Line~\ref{alg:line:10}). We employ a modified inverse distance weighting procedure inspired from~\cite{10.1145/800186.810616} to estimate the likelihood that our observed $A^*$ is scenario $s$, generating a probability likelihood state vector for all $S$ (Line~\ref{alg:line:11}). We calculate the swarm markers state ($\vert\vert\mathcal{M}^\Pi\vert\vert_2$, Line~\ref{alg:line:13}) using the same technique for $\mathcal{M^\pi}$ and probabilistically classify the marker state as a particular swarm type (Line~\ref{alg:line:14}). We combine the outputs of this and the previous method into a single ensemble to classify the most likely observed swarm scenario ($\widehat{\Pi}$). We select $\widehat{\Pi}^*$ by selecting the most likely scenario overall context windows observed from $t=0$ to the current observation period, considering probabilistic variability. In practical terms, we select the scenario that returns the maximum likelihood as $\argmax(\text{mean}(\widehat{\Pi}) - \text{var}(\widehat{\Pi}))$ (Line~\ref{alg:line:15}).

We conduct pair-wise statistical tests to determine the significance of $\widehat{\Pi}^*$, comparing to all other $\widehat{\Pi}$ (Line~\ref{alg:line:16}). For the case where $\widehat{\Pi}^*$ is significant (i.e. has statistically higher likelihood than all $\widehat{\Pi}$) and the tactic pair (TP, $\mathfrak{T}$) currently employed is significant (Line~\ref{alg:line:17}), then we parameterise $\beta$ with $\mathfrak{T}$ (Line~\ref{alg:line:18}). In the case where $\mathfrak{T}$ is not significant, then we select a new $\mathfrak{T}$ from $\Sigma$ library that maximises the likelihood of mission success, for the given scenario (Line~\ref{alg:line:20}). In the case where $\widehat{\Pi}^*$ is not significant (i.e. does not have a higher likelihood than $\widehat{\Pi}$), we select the highest likelihood scenario type as a two-class problem between heterogeneous or homogeneous Line~\ref{alg:line:23} and parameterise $\mathfrak{T}$ in either of these forms (i.e. not for a specific scenario but a class of scenarios, Line~\ref{alg:line:24}). Our final tasks are to determine $\sigma^{c_2}$ (Line~\ref{alg:line:26}) and $\sigma^{c_3}$ (Line~\ref{alg:line:27}) from $\widehat{\Pi}^*$ and $\mathfrak{T}$, then parameterise $\beta$ for the next context period to influence the swarm (Line~\ref{alg:line:28}).

\begin{algorithm}[h!]
    \caption{Context-Aware Decision Shepherding Agent ($\omega, \tau$)}\label{Algo:ContextAnalytics}
    \begin{algorithmic}[1]
        \State Set context window length $\omega$ and context window overlap $\tau$                         \label{alg:line:1}
        \While{$\vert\vert P_\Pi - P_G \vert\vert > R_G$}                                                   \label{alg:line:2}
            \State Observe $P^{\Delta\omega}_{\pi_i}\quad\forall~\pi_i \in \Pi$                             \label{alg:line:3}
            \For{$\pi_i \in \Pi$}                                                                           \label{alg:line:4}
                \State Calculate $\mathcal{M}^{\pi_i}$                                                      \label{alg:line:5}
                \State Classify $\widehat{\pi_i}: \mathcal{M}^{\pi_i} \rightarrow A$                        \label{alg:line:6}
            \EndFor                                                                                         \label{alg:line:7}
            \State Summarise $A^* = \text{mean}(A)$ \Comment{$\forall\quad A$}                              \label{alg:line:8}
            \For{$s\in S$}                                                                                  \label{alg:line:9}
                \State Calculate $\vert\vert s - A^*\vert\vert_2$                                           \label{alg:line:10}
                \State Calculate inverse weighting distance to obtain $Pr(A^*)\rightarrow s$                \label{alg:line:11}
            \EndFor                                                                                         \label{alg:line:12}
            \State Calculate $\vert\vert\mathcal{M}^{\Pi}\vert\vert_2$                                      \label{alg:line:13}
            \State Classify $\widehat{\Pi}: \mathcal{M}^{\Pi} \rightarrow S$                                \label{alg:line:14}
            \State $\widehat{\Pi}^* = \argmax(\text{mean}(\widehat{\Pi}) - \text{var}(\widehat{\Pi}))$      \label{alg:line:15}
            \If{$\widehat{\Pi}^*$ significant}                                                              \label{alg:line:16}
                \If{$\mathfrak{T}^t$ significant}                                                           \label{alg:line:17}
                    \State Continue $\mathfrak{T}^t$                                                        \label{alg:line:18}
                \Else                                                                                       \label{alg:line:19}
                    \State Select $\argmax(\text{TP}\rightarrow s^*)$                                       \label{alg:line:20}
                \EndIf                                                                                      \label{alg:line:21}
            \Else                                                                                           \label{alg:line:22}
                \State Classify $\widehat{\Pi}^* \rightarrow S = \argmax(s_{He},  s_{Ho})$                  \label{alg:line:23}
                \State Set $\mathfrak{T}$ to $S^*$                                                          \label{alg:line:24}
            \EndIf                                                                                          \label{alg:line:25}
            \State Calculate $\sigma^{c_2} \sim \widehat{\Pi}^*, \mathfrak{T}$                              \label{alg:line:26}
            \State Calculate $\sigma^{c_3} \sim \widehat{\Pi}^*, \mathfrak{T}$                              \label{alg:line:27}
            \State Parameterise $\beta$ = $<\sigma^{c_2}, \sigma^{c_3}, \mathfrak{T}>$                      \label{alg:line:28}
        \EndWhile                                                                                           \label{alg:line:29}
    \end{algorithmic}
\end{algorithm}

\subsection{Multi-Agent Reasoning for Swarm Control}

\cite{Singh2019:Modulation} state that \enquote{Smart $\pi$ and $\beta$ agents in a real-world robotic setting need to comprehend their environment and make appropriate decisions on how to modulate their speed in response to the state and changes of their internal and external contextual information}. Our system architecture aims to solve a portion of this challenge and consists of three agent types: the context-aware AI decision support system agent, cognitive agent, and shepherding agent. Each agent is independent and specialises in a single task; we have discussed the context-awareness system and now describe cognitive and shepherding agent functions.

\begin{itemize}

\item \textbf{Cognitive Agent}

In the architecture, the cognitive agent determines which behaviour from the context-aware agent's parameters should be conducted, selecting between $\sigma_1$ and $\sigma_2$, consistent with prior shepherding models. The function of the cognitive agent is to calculate $f(N)$ to select a behaviour, $\sigma^*$, to be conducted, determined to be either $\sigma_1$ or $\sigma_2$. The cognitive agent makes this decision with the parameterised behaviours provided by the context agent and the positions of all swarm agents in the environment.

\item \textbf{Shepherding Agent}

The shepherding agent is the actuator that delivers force vectors to influence the swarm to move to the goal location. The shepherding agent decides on a behaviour $\sigma^*$ and how to conduct it, calculating the positioning point ($P^t_{\beta_\sigma}$) for the behaviour. The function of the shepherding agent is to calculate the behaviour position point and execute a force vector to influence the swarm based on the behaviour selected and the configuration of the swarm.

\end{itemize}

\section{S2AI: Sensor to Actionable Information}\label{sec:S2AI}

This section aims to examine the effectiveness of different \emph{collect} and \emph{drive} behaviours in \cite{Strombom:2014} on the success of shepherding. We introduce two points of departure from the original model: the modulation of the control behaviours available to the shepherding agent and the inclusion of heterogeneous swarm agent parameterised weights. Our intent in systematically varying these aspects of the model is to investigate the performance of distinct combinations of autonomous control behaviours (collect and drive actions), across different homogeneous and heterogeneous swarm scenarios.

\subsection{Shepherding Agent Behaviours}

A single instantiation of each behaviour is usually presented in the majority of shepherding-based models, for instance~\cite{1308924, Strombom:2014}, often activated through a linear rule-based switching mechanism that operates on low-level positional information of each of the swarm agents. This study introduces an expanded set of five collect and five drive behaviours. The driving position for the drive behaviour ($\sigma_1$) as given in \cite{Strombom:2014} is calculated by the function
\begin{equation}
    P^t_{\beta\sigma_1} = \Lambda^t_{\beta} + f(N) \cdot \dfrac{\Lambda^t_{\beta} - P^t_G}{\vert\vert \Lambda^t_{\beta} - P^t_G \vert\vert},
\end{equation}
where $f(N) = R_{\pi\pi}N^{\sfrac{2}{3}}$. $P^t_{\beta\sigma_1}$ is the position for agent $\beta$ to execute behaviour $\sigma_1$ at time $t$; $\Lambda^t_\beta$ is the local centre of mass for the swarm to be controlled by agent $\beta$ at $t$; and $\sfrac{\Lambda^t_{\beta} - P^t_G}{\vert\vert \Lambda^t_{\beta} - P^t_G \vert\vert}$ the direction in which the driving point needs to be oriented. The driving behaviour is triggered if the following inequality, signifying that the swarm is well-grouped, is satisfied
\begin{equation}
    \forall~\pi_i\in\Omega^t_{\beta\pi}, \vert\vert \Lambda_{\beta^t} - P^t_{\pi} \vert\vert \leq f(N),
\end{equation}
where $\Omega^t_{\beta\pi}$ is the set of $\pi$ agents $\beta$ operates on. We summarise the algorithm by \cite{Strombom:2014} in Algorithm~\ref{Algo:ClassicReactive}.

\begin{algorithm}[h!]
    \caption{Reactive Shepherding Agent}\label{Algo:ClassicReactive}
    \begin{algorithmic}[1]
        \While{$\vert\vert P_\Pi - P_G \vert\vert > R_G$} \Comment{$R_G$ is the goal radius}
            \State Observe $P^t_{\pi_i}~\forall~i$
            \State Set $\Omega^t_{\beta\pi} = \min \{ \Omega^t,\Omega_{\beta\pi} \}$ of $\pi_i$ agents to $\beta$
            \State Set $\Lambda^t_\beta$ as the centre of mass
            \State Calculate $f(N) = R_{\pi\pi}N^{\sfrac{2}{3}}$
            \If{$\forall~\pi_i\in\Omega^t_{\beta\pi}, \vert\vert \Lambda_{\beta^t} - P^t_{\pi} \vert\vert \leq f(N)$}
                \State Conduct driving behaviour, $\sigma_1$
            \Else $~\exists~\pi^*_i\in\Omega_{\beta\pi}^t, \argmax_i(\vert\vert \Lambda^t_{\beta} - P^t_{\pi_i} \vert\vert > f(N))$
                \State Conduct collecting behaviour, $\sigma_2$
            \EndIf
        \EndWhile
    \end{algorithmic}
\end{algorithm}
 
Note that in Algorithm~\ref{Algo:ClassicReactive}, the inequality for $f(N)$ is calculated at every period. We modify the term $f(N)$ by including a threshold $\mathcal{L}$ that modulates $N$. This serves as a constraint on the number of agents $\beta$ operates on to determine if $\Pi$ is clustered for the purpose of driving. The resulting inequality is now given as
\begin{equation}\label{eqn:drive}
    \forall~\pi_i\in\Omega^t_{\beta\pi}, \vert\vert \Lambda_{\beta^t} - P^t_{\pi} \vert\vert \leq f(\mathcal{L}N),
\end{equation}
where we set $\mathcal{L} = \{1.00, 0.75, 0.50, 0.25, \sfrac{1}{N} \}$ to generate the additional four behaviours. We abbreviate each behaviour as drive-100 (\texttt{D100}, $\mathcal{L} = 1.00$), drive-75 (\texttt{D75}, $\mathcal{L} = 0.75$), drive-50 (\texttt{D50}, $\mathcal{L} = 0.50$), and drive-25 (\texttt{D25}, $\mathcal{L} = 0.25$). The final drive behaviour is a special case we call drive-one (\texttt{D1N}, $\mathcal{L} = \sfrac{1}{N}$), being the selection of a single agent for driving. The collecting point behaviour ($\sigma_2$) can be given by the function
\begin{equation}
    P^t_{\beta\sigma_2} = P^t_{\pi^*_i} + R_{\pi\pi}\cdot\dfrac{P^t_{\pi^*_i} - \Lambda^t_{\beta}}{\vert\vert P^t_{\pi^*_i} - \Lambda^t_{\beta} \vert\vert}.
\end{equation}
The attention of the swarm agent is focused on the agent furthest from the swarm, activated by the inequality
\begin{equation}
    \exists~\pi^*_i\in\Omega_{\beta\pi}^t, \argmax_i(\vert\vert \Lambda^t_{\beta} - P^t_{\pi_i} \vert\vert > f(N)).
\end{equation}

We introduce additional collect reference points to generate four different behaviours, modifying the selection of a single $\pi_i$ to collect. The collecting behaviours available to the shepherding agent include
\begin{itemize}
    \item select the $\pi_i$ closest to the shepherding agent (\texttt{C2D}, closest-to-dog),
    \item select the $\pi_i$ closest to the $\Pi$ (\texttt{C2H}, closest-to-herd centre of mass),
    \item select the $\pi_i$ furthest to $\beta$ (\texttt{F2D}, furthest-to-dog),
    \item select the $\pi_i$ furthest to $P_G$ (\texttt{F2G}, furthest-to-goal), or
    \item select the $\pi_i$ furthest to $\Pi$ (\texttt{F2H}, furthest-to-herd centre of mass).
\end{itemize}
Algorithm~\ref{Algo:MultiSigmaReactive} summarises changes to Algorithm~\ref{Algo:ClassicReactive}. The key changes between these algorithms include
\begin{enumerate}
    \item initialising $\sigma_1$ to set $\mathcal{L}\in [0,1]$, and
    \item initialising $\sigma_2$ to one of the five focus points (as outlined in the previous paragraph).
\end{enumerate}

\begin{algorithm}[h!]
    \caption{Multiple Behaviour Reactive Shepherding Agent ($\sigma_1^i, \sigma_2^j$)}\label{Algo:MultiSigmaReactive}
    \begin{algorithmic}[1]
        \While{$\vert\vert P_\Pi - P_G \vert\vert > R_G$}
            \State Observe $P^t_{\pi_i}~\forall~i$
            \State Set $\Omega^t_{\beta\pi} = \min \{ \Omega^t,\Omega_{\beta\pi} \} $ of $\pi_i$ agents to $\beta$
            \State Set $\Lambda^t_\beta$ as the centre of mass
            \State Calculate $f(N) = R_{\pi\pi}N^{\sfrac{2}{3}}$
            \If {$\forall~\pi_i\in\Omega^t_{\beta\pi}, \vert\vert \Lambda_{\beta^t - P^t_{\pi}} \vert\vert \leq f(\mathcal{L}N)$}
                \State Conduct driving behaviour, $\sigma_1$
            \Else $\ $ based on the collect reference point,
                \State Conduct collecting behaviour, $\sigma_2$
            \EndIf
        \EndWhile
    \end{algorithmic}
\end{algorithm}

We parameterise the primitive behaviours of collect and drive for the control agent, implemented in a behaviour library and summarised in Table~\ref{table:AgentTactics}.

    \begin{table}[h!]
    \centering
    \def\arraystretch{1.5}
    \resizebox{0.9\textwidth}{!}{%
    \begin{tabular}{@{}cccl@{}}
        \toprule
        \textbf{Behaviour} & \textbf{Type} & \textbf{Nomenclature} & \textbf{Description}                         \\ \midrule
        \texttt{D1N}        & Drive         & $\sigma_1^{D1N}$          & Drive exactly one agent of the herd         \\
        \texttt{D25}        & Drive         & $\sigma_1^{D25}$          & Drive at least $\frac{1}{4}$ of the herd          \\
        \texttt{D50}        & Drive         & $\sigma_1^{D50}$          & Drive at least $\frac{1}{2}$ of the herd          \\
        \texttt{D75}        & Drive         & $\sigma_1^{D75}$          & Drive at least $\frac{3}{4}$ of the herd          \\
        \texttt{D100}       & Drive         & $\sigma_1^{D100}$         & Drive all of the herd                             \\
        \texttt{C2D}        & Collect       & $\sigma_2^{C2D}$          & Closest swarm agent to shepherding agent          \\
        \texttt{C2H}        & Collect       & $\sigma_2^{C2H}$          & Closest swarm agent to the herd centre of mass    \\
        \texttt{F2D}        & Collect       & $\sigma_2^{F2D}$          & Furthest swarm agent to shepherding agent         \\
        \texttt{F2G}        & Collect       & $\sigma_2^{F2G}$          & Furthest swarm agent to the goal                  \\
        \texttt{F2H}        & Collect       & $\sigma_2^{F2H}$          & Furthest swarm agent to the herd centre of mass   \\
        \bottomrule
    \end{tabular}
    }
    \caption{Summary of control agent behaviours parameterised in this study.}
    \label{table:AgentTactics}
\end{table}

Our shepherding model instantiates a one collect and one drive behaviour employed by $\beta$ to control $\Pi$. We denote a pair of behaviour as a \emph{tactic pair} (TP); that is, $TP = \{ \sigma_1, \sigma_2 \}$; in the remainder of this paper, we use a unique TP label instead of behaviour combinations using such as $\{ \sigma_1^{D100}, \sigma_2^{F2H} \}$. Our objective with all TPs is to evaluate the combinations of collect and drive behaviours, consisting of 25 TPs (TP$_{1,\dots,25}$). Table~\ref{table:TacticPairs} introduces the parameterisations of collect and drive behaviours in this study.

\begin{table}[h!]
    \begin{subtable}[h]{0.475\textwidth}
        \centering
        \def\arraystretch{1.25}
        \resizebox{\textwidth}{!}{%
        \begin{tabular}{ccccccccccccc}
            \toprule
                \textbf{Tactic Pair} & \textbf{Notation} & \textbf{Behaviours} \\ \midrule
                 TP$_1$    & $\{\sigma_1^{D100}, \sigma_2^{C2D}\}$ & \texttt{D100, C2D} \\
                 TP$_2$    & $\{\sigma_1^{D100}, \sigma_2^{C2H}\}$ & \texttt{D100, C2H} \\
                 TP$_3$    & $\{\sigma_1^{D100}, \sigma_2^{F2D}\}$ & \texttt{D100, F2D} \\
                 TP$_4$    & $\{\sigma_1^{D100}, \sigma_2^{F2G}\}$ & \texttt{D100, F2G} \\
                 TP$_5$$\ast$& $\{\sigma_1^{D100}, \sigma_2^{F2H}\}$&\texttt{D100, F2H} \\
                 TP$_6$    & $\{\sigma_1^{D50}, \sigma_2^{C2D}\}$ & \texttt{D50, C2D}   \\
                 TP$_7$    & $\{\sigma_1^{D50}, \sigma_2^{C2H}\}$ & \texttt{D50, C2H}   \\
                 TP$_8$    & $\{\sigma_1^{D50}, \sigma_2^{F2D}\}$ & \texttt{D50, F2D}   \\
                 TP$_9$    & $\{\sigma_1^{D50}, \sigma_2^{F2G}\}$ & \texttt{D50, F2G}   \\
                 TP$_{10}$ & $\{\sigma_1^{D50}, \sigma_2^{F2H}\}$ & \texttt{D50, F2H}   \\
                 TP$_{11}$ & $\{\sigma_1^{D1N}, \sigma_2^{C2D}\}$ & \texttt{D1N, C2D}   \\
                 TP$_{12}$ & $\{\sigma_1^{D1N}, \sigma_2^{C2H}\}$ & \texttt{D1N, C2H}   \\
            \bottomrule
        \end{tabular}
        }
    \end{subtable}
    \hfill
    \begin{subtable}[h]{0.445\textwidth}
        \centering
        \def\arraystretch{1.25}
        \resizebox{\textwidth}{!}{%
        \begin{tabular}{ccccccccccccc}
            \toprule
                \textbf{Tactic Pair} & \textbf{Notation} & \textbf{Behaviours} \\ \midrule
                 TP$_{13}$ & $\{\sigma_1^{D1N}, \sigma_2^{F2D}\}$ & \texttt{D1N, F2D} \\
                 TP$_{14}$ & $\{\sigma_1^{D1N}, \sigma_2^{F2G}\}$ & \texttt{D1N, F2G} \\
                 TP$_{15}$ & $\{\sigma_1^{D1N}, \sigma_2^{F2H}\}$ & \texttt{D1N, F2H} \\
                 TP$_{16}$ & $\{\sigma_1^{D25}, \sigma_2^{C2D}\}$ & \texttt{D25, C2D} \\
                 TP$_{17}$ & $\{\sigma_1^{D25}, \sigma_2^{C2H}\}$ & \texttt{D25, C2H} \\
                 TP$_{18}$ & $\{\sigma_1^{D25}, \sigma_2^{F2D}\}$ & \texttt{D25, F2D} \\
                 TP$_{19}$ & $\{\sigma_1^{D25}, \sigma_2^{F2G}\}$ & \texttt{D25, F2G} \\
                 TP$_{20}$ & $\{\sigma_1^{D25}, \sigma_2^{F2H}\}$ & \texttt{D25, F2H} \\
                 TP$_{21}$ & $\{\sigma_1^{D75}, \sigma_2^{C2D}\}$ & \texttt{D75, C2D} \\
                 TP$_{22}$ & $\{\sigma_1^{D75}, \sigma_2^{C2H}\}$ & \texttt{D75, C2H} \\
                 TP$_{23}$ & $\{\sigma_1^{D75}, \sigma_2^{F2D}\}$ & \texttt{D75, F2D} \\
                 TP$_{24}$ & $\{\sigma_1^{D75}, \sigma_2^{F2G}\}$ & \texttt{D75, F2G} \\
                 TP$_{25}$ & $\{\sigma_1^{D75}, \sigma_2^{F2H}\}$ & \texttt{D75, F2H} \\
            \bottomrule
        \end{tabular}
        }
    \end{subtable}
    \caption{The set of 25 tactic pairs used in this study systematically explores 5 variations of each $\sigma_1$ and $\sigma_2$. TP$_5$ (denoted with $\ast$) represents the behaviour introduced by \cite{Strombom:2014}.}
    \label{table:TacticPairs}
\end{table}

\subsection{Swarm Heterogeneity}
We vary 3 weights of this $\pi$-agents and the speed differential to introduce 6 additional $\pi$-types, parameterised as presented in \cite{Hepworth2022:SwarmAnalytics} (see Table~4 from \cite{Hepworth2022:SwarmAnalytics} for further detail). Specifically, we vary
\begin{itemize}
    \item $W_{\pi\Lambda}$: strength of attraction for a $\pi$ to their local centre of mass, $\Lambda$.
    \item $W_{\pi\pi}$: strength of repulsion for a $\pi$ to another $\pi$.
    \item $W_{\pi\beta}$: strength of repulsion for a $\pi$ to the shepherding agent $\beta$.
    \item $\dfrac{s_\pi}{s_\beta}$: speed differential between a $\pi$ agent and $\beta$.
\end{itemize}
The 7 $\pi_i$ agent types and parameterised weights in this study are given in Table~\ref{table:AgentCharacteristics}, as originally presented in \cite{Hepworth2022:SwarmAnalytics}. The parameterisation of agent A7 is as initially presented by \cite{Strombom:2014}.

    \begin{table}[h!]
        \centering
        \resizebox{0.65\textwidth}{!}{%
        \begin{tabular}{ccccc}
            \toprule
            \textbf{Agent State} & \textbf{$W_{\pi\Lambda}$} & \textbf{$W_{\pi\pi}$} & \textbf{$W_{\pi\beta}$} & \textbf{$\sfrac{s_\pi}{s_\beta}$} \\ \midrule
             A1 & 0.50 & 2.00 & 0.50 & 1.00 \\
             A2 & 1.50 & 2.00 & 0.50 & 0.50 \\
             A3 & 0.50 & 3.00 & 1.00 & 0.67 \\
             A4 & 0.50 & 2.00 & 1.90 & 0.67 \\
             A5 & 1.05 & 3.00 & 1.00 & 0.67 \\
             A6 & 1.05 & 1.50 & 1.00 & 0.50 \\
             A7 & 1.05 & 2.00 & 1.00 & 0.67 \\
             \bottomrule
        \end{tabular}
        }
        \caption{Summary of agent state vector weight parameterisations.}
        \label{table:AgentCharacteristics}
    \end{table}

The $\pi$-type agents are allocated to different swarms as the constituent members of 11 scenarios, consisting of four heterogeneous and seven homogeneous swarms, presented in Table~5 of \cite{Hepworth2022:SwarmAnalytics} and summarised here in Table~\ref{table:SwarmScenarios}. \cite{Strombom:2014} employ swarm exclusively $S5$ in their study, a homogeneous scenario with agent type A7. We include an additional 6 homogeneous swarms scenarios ($S5, S6, S7, S8, S9, S10, S11$), one for each agent type in Table~\ref{table:AgentCharacteristics}. We also parameterise four heterogeneous swarm scenarios ($S1, S2, S3, S4$), as first presented in~\cite{Hepworth2022:SwarmAnalytics}.

        \begin{table}[h!]
        \centering
        \resizebox{0.8\textwidth}{!}{%
        \begin{tabular}{cccccccc}
            \toprule
            \textbf{Scenario State} & \textbf{A1} & \textbf{A2} & \textbf{A3} & \textbf{A4} & \textbf{A5} & \textbf{A6} & \textbf{A7}  \\ \midrule
             $S1$  & 0.20 &      &      &      &      &      & 0.80 \\
             $S2$  &      & 0.20 & 0.20 &      &      & 0.20 & 0.40 \\
             $S3$  &      &      &      & 0.80 &      &      & 0.20 \\
             $S4$  & 0.20 &      &      &      & 0.20 &      & 0.60 \\
             $S5$  &      &      &      &      &      &      & 1.00 \\
             $S6$  & 1.00 &      &      &      &      &      &      \\
             $S7$  &      & 1.00 &      &      &      &      &      \\
             $S8$  &      &      & 1.00 &      &      &      &      \\
             $S9$  &      &      &      & 1.00 &      &      &      \\
             $S10$ &      &      &      &      & 1.00 &      &      \\
             $S11$ &      &      &      &      &      & 1.00 &      \\
             \bottomrule
        \end{tabular}
        }
        \caption{Summary of swarm scenarios used in this study, as originally presented in \cite{Hepworth2022:SwarmAnalytics}.}
        \label{table:SwarmScenarios}
    \end{table}

\subsection{Experimental Design and Analysis}

Our experimental design aims to evaluate each TP\textquoteright s performance in both homogeneous and heterogeneous swarm settings. We assess the performance of each TP for a particular scenario 30 times, resulting in a total of 8,250 trials (30 trials for 25 TPs in 11 scenarios); 20 $\pi$ agents and 1 $\beta$ agent was used similar to~\cite{9659082}. In addition, we collect a range of summary statistics on TP performance, summarised by the metrics presented in Table~\ref{table:metrics}.

\cite{rizk2019a} discuss the development of standardised evaluation metrics, highlighting that \enquote{the evaluation criteria include specific domain performance metrics and domain invariant criteria}, offering examples such as spatiotemporal, complexity, load, fairness, communication, robustness, scalability and resource-based metrics. Selecting appropriate evaluation metrics depends on the model understanding desired, which can include top-level characteristics such as autonomy, complexity, adaptability, concurrency, distribution and communication~\cite{10.1145/375735.376473}.

Various methods have been proposed to evaluate the performance of swarm control methods. These methods often compare the performance of one control approach to another or to understand the sensitivity of particular parameter settings; for instance, see variations of \cite{Strombom:2014} such as \cite{9256255, Zhang2022:Herding, Singh2019:Modulation}. However, measures to evaluate the inclusion of cognitive capabilities as outlined by \cite{Long2020:Comprehensive} are yet to be widely established, highlighting the requirement for an evaluation approach to account for the contribution of new capabilities with the agent.

To address this, we propose four new metrics specific for swarm shepherding to evaluate the system and include two metrics common for performance analysis. Our metrics consider three aspects of task performance and two aspects of control agent adaptability, quantifying the impact of context on the agent. Table~\ref{table:metrics} summarises the metrics we use to evaluate tactic pairs. Our three selected task performance metrics are well established for the analysis of swarm shepherding systems; see, for example~\cite{9256255, Strombom:2014, 9659082}, who further discuss measures of this class.

The first metric is Mission Success, indicating if the overall goal was achieved. Mission Completion Rate is the second metric and measures the effectiveness of a selected control strategy for a mission, indicating proximity to the goal location from the starting position of the swarm. The third metric is Mission Speed, measuring the effectiveness of a selected control strategy in terms of the average speed of the control agent to move the swarm throughout the mission.

As well as the three task performance measures discussed above, we present three measures of stability based on the control agent adaptability metrics discussed. The two adaptability aspects include decision changes and separated agents. Exploring the number of decision changes is an established metric, for example, to bound the computations of an agent~\cite{10.1007/978-3-642-28499-1_4} or investigate the effect of environmental conditions on decision frequency~\cite{Mills2015:Behaviour}. In our setting, we use this metric to quantify the impact a swarm context has on the number of decision changes a control agent needs to make. In addition, quantifying spatially separated agents can help understand collective decision processes and their impacts on a group, for example, the evolution of distinct population groups or agent couplings~\cite{Mills2015:Behaviour}. Finally, we use the number of separated agents to measure the impact of the control agent\textquoteright s influence on the stability of the swarm.

The first stability metric is Mission Decision Stability, highlighting if a small change in the system will result in a large decision space change. We define the function $\chi_1$ as
\begin{equation}
    \chi_1(\sigma(t), \sigma(t+1)) = \begin{cases}
                                            1, & \text{if}~\sigma(t)\neq\sigma(t+1) \\
                                            0, & \text{otherwise} \end{cases},
\end{equation}
which provides a count for the number of changes of $\sigma$ behaviours the shepherding control agent conducts in successive time steps. For example, if the behaviour at $t$ is $\sigma_2$ and subsequently at $t+1$ is $\sigma_1$, then $\chi$ returns the value of 1. If $\sigma$ does not change for $t+1$, then 0 is returned. The second stability metric is Decision Swarm Stability, which informs on the relationship between the number of separated swarm agents and the number of decisions made by the control agent, highlighting if swarm instability (fracture) is related to control decision changes. We define the function $\chi_2$ as
\begin{equation}
    \chi_2(\pi_i, \Pi^*) = \begin{cases}
                                            1, & \text{if}~ \pi_i \notin \Pi^* \\
                                            0, & \text{otherwise} \end{cases},
\end{equation}
where $\Pi^*$ is the cluster size with the largest number of $\pi$ agents, calculated using the k-means algorithm. We evaluate the position of each $\pi_i$ to determine if a swarm agent is separated from this cluster; if the agent is assessed to be separated, we return the value 1 and 0 otherwise.

The final stability metric is Mission Swarm Stability, indicating if the swarm control agent can overcome instability in the swarm, such as $\pi_i$ agents with considerable repulsion strengths for $W_{\pi\pi}$ or $W_{\pi\beta}$.

\begin{table}[h!]
    \centering
    \def\arraystretch{3.5}
    \resizebox{\textwidth}{!}{%
    \begin{tabular}{llL{15em}}
        \toprule
        \textbf{Metric}                 & \textbf{Equation}                                                                                                                                                  & \textbf{Interpretation} \\ \midrule
             Mission Success            & $MS = \begin{cases}
                                                1, & \text{if}\ P^T_{\Pi} - P^T_G < R_G \\
                                                0, & \text{otherwise} \end{cases}$                                                                                                                          & Effectiveness metric for the overall mission success. Over multiple trials, the objective is to maximise the number of successes. \\
        Mission Completion Rate         & $MCR = \dfrac{\vert\vert P^T_\Pi - P^T_G \vert\vert}{\vert\vert P^T_\Pi - P^{t=0}_\Pi \vert\vert}$                                                                & Efficiency metric for the shepherding agent\textquoteright s control strategy. A lower mission completion rate indicates increased performance for this metric. \\
        Mission Speed                   & $MSp = \dfrac{\vert\vert P^T_\Pi - {P^{t=0}_\Pi} \vert\vert}{\Delta t}$                                                                                            & Efficiency metric measuring the pace of mission execution. Higher mission speed indicates increased performance. \\
        Mission Decision Stability & $MDS = \dfrac{MS}{1+\sum_{j=1}^T{\chi_1(\sigma(t), \sigma(t+1))}}$                                                                                                & Sensitivity metric informing on the mission impact to changes in the number of control decisions. Higher mission decision stability indicates increased performance for this metric. \\
        Decision Swarm Stability        & $DSS = \dfrac{\sfrac{1}{\vert\Pi\vert}\sum_{j=1}^{j=T}\sum_{i=1}^{\vert\Pi\vert}{\chi_2(\pi_i, \Pi^*)_j}}{1+\sum_{j=1}^T{\chi_1(\sigma(t), \sigma(t+1))}}$        & Sensitivity metric informing if a control strategy could be considered susceptible to perturbations in change. Lower decision swarm stability indicates increased performance for this metric.\\
        Mission Swarm Stability    & $MSS = \dfrac{MS}{1+\sfrac{1}{\vert\Pi\vert}\sum_{j=1}^{j=T}\sum_{i=1}^{\vert\Pi\vert}{\chi_2(\pi_i, \Pi^*)_j}}$                                                  & Sensitivity metric informing on the mission impact to changes in swarm stability. Higher mission swarm stability indicates increased performance for this metric.\\
        \bottomrule
    \end{tabular}
    }
    \caption{Metrics used in this study and interpretations for tactic pair assessments. A single metric is reported for each simulation trial as a scenario summary, calculated at each simulation step.}
    \label{table:metrics}
\end{table}

We evaluate and compare the performance of TPs in two ways. The first is a comparative analysis of TP$_5$ as a baseline representation of the algorithm in \cite{Strombom:2014} to the TP with the highest performance for that scenario, what we label as the \emph{Best TP}. The Best TP is determined for each metric and scenario by calculating the mean and selecting the TP with the highest mean for the particular setting. We then conduct the $t$-test to test for statistical significance of the difference between the results of TP$_5$ and the best TP, with the null hypothesis being an absence of difference between the results. Table~\ref{table:stats} compares performance distinct for each metric and scenario to that of TP$_5$. It can be seen that $TP_5$ outperforms another TP across most scenarios and metrics. This demonstrates that using a single TP for homogeneous and heterogeneous swarms parameterised per that of \cite{Strombom:2014} may not be suitable in all circumstances for swarm control, invariant of the metric used to compare performance.

\begin{table}[h!]
    \begin{subtable}[h]{\textwidth}
        \centering
        \resizebox{\textwidth}{!}{%
        \begin{tabular}{ccccccccccccc}
            \toprule
            \multicolumn{1}{c}{} &
              \multicolumn{2}{c}{\textbf{MS}} &
              \multicolumn{2}{c}{\textbf{MDS}} &
              \multicolumn{2}{c}{\textbf{DSS}} \\
             & \textbf{Best TP} & \textbf{TP$_5$} & \textbf{Best TP} & \textbf{TP$_5$} & \textbf{Best TP} & \textbf{TP$_5$} \\
             \midrule
            \multicolumn{1}{c}{$\widehat{\mathbf{S}}$} & \textbf{0.68 $\pm$ 0.47} (TP$_{4}$) & 0.64 $\pm$ 0.48          & \textbf{0.27 $\pm$ 0.23} (TP$_{4}$) & 0.13 $\pm$ 0.19 & \textbf{0.29 $\pm$ 0.87} (TP$_{25}$) & 1.14 $\pm$ 1.81 \\
            \midrule
            \textbf{S1}                     & \textbf{0.90 $\pm$ 0.31} (TP$_{4}$) & 0.73 $\pm$ 0.45          & \textbf{0.37 $\pm$ 0.18} (TP$_{4}$)  & 0.07 $\pm$ 0.13       & \textbf{0.18 $\pm$ 0.26} (TP$_{25}$) & 0.83 $\pm$ 1.39       \\
            \textbf{S2}                     & \textbf{0.87 $\pm$ 0.35} (TP$_{4}$) & 0.33 $\pm$ 0.48$\ast$    & \textbf{0.33 $\pm$ 0.19} (TP$_{4}$)  & 0.10 $\pm$ 0.19$\ast$ & \textbf{0.03 $\pm$ 0.02} (TP$_{25}$) & 1.65 $\pm$ 2.20       \\
            \textbf{S3}                     & \textbf{0.50 $\pm$ 0.51} (TP$_{5}$) & \textbf{0.50 $\pm$ 0.51} & \textbf{0.30 $\pm$ 0.47} (TP$_{15}$) & 0.09 $\pm$ 0.13       & \textbf{0.06 $\pm$ 0.05} (TP$_{19}$) & 1.74 $\pm$ 2.42$\ast$ \\
            \textbf{S4}                     & \textbf{0.90 $\pm$ 0.31} (TP$_{9}$) & 0.62 $\pm$ 0.49$\ast$    & \textbf{0.38 $\pm$ 0.19} (TP$_{4}$)  & 0.06 $\pm$ 0.12$\ast$ & \textbf{0.10 $\pm$ 0.14} (TP$_{25}$) & 1.02 $\pm$ 1.81       \\
            \textbf{S5}                     & \textbf{1.00 $\pm$ 0.00} (TP$_{5}$) & \textbf{1.00 $\pm$ 0.00} & \textbf{0.32 $\pm$ 0.23} (TP$_{4}$)  & 0.06 $\pm$ 0.05$\ast$ & \textbf{0.14 $\pm$ 0.11} (TP$_{25}$) & 0.41 $\pm$ 0.42       \\
            \textbf{S6}                     & \textbf{0.83 $\pm$ 0.38} (TP$_{4}$) & 0.37 $\pm$ 0.49$\ast$    & \textbf{0.32 $\pm$ 0.19} (TP$_{4}$)  & 0.04 $\pm$ 0.11$\ast$ & \textbf{0.06 $\pm$ 0.06} (TP$_{25}$) & 0.45 $\pm$ 0.86       \\
            \textbf{S7}                     & \textbf{1.00 $\pm$ 0.00} (TP$_{4}$) & 1.00 $\pm$ 0.00          & \textbf{0.97 $\pm$ 0.18} (TP$_{15}$) & 0.50 $\pm$ 0.00       & \textbf{0.41 $\pm$ 0.29} (TP$_{16}$) & 2.03 $\pm$ 0.43       \\
            \textbf{S8}                     & \textbf{0.30 $\pm$ 0.47} (TP$_{4}$) & 0.07 $\pm$ 0.25$\ast$    & \textbf{0.12 $\pm$ 0.20} (TP$_{4}$)  & 0.02 $\pm$ 0.06$\ast$ & \textbf{0.03 $\pm$ 0.02} (TP$_{10}$) & 0.25 $\pm$ 0.72       \\
            \textbf{S9}                     & \textbf{0.81 $\pm$ 0.40} (TP$_{5}$) & \textbf{0.81 $\pm$ 0.40} & \textbf{0.43 $\pm$ 0.50} (TP$_{11}$) & 0.11 $\pm$ 0.12$\ast$ & \textbf{0.06 $\pm$ 0.04} (TP$_{19}$) & 2.21 $\pm$ 3.44$\ast$ \\
            \textbf{S10}                    & \textbf{1.00 $\pm$ 0.00} (TP$_{4}$) & 0.37 $\pm$ 0.49$\ast$    & \textbf{0.40 $\pm$ 0.14} (TP$_{4}$)  & 0.02 $\pm$ 0.05$\ast$ & \textbf{0.11 $\pm$ 0.08} (TP$_{25}$) & 0.23 $\pm$ 0.37$\ast$ \\
            \textbf{S11}                    & \textbf{0.97 $\pm$ 0.18} (TP$_{25}$)& 0.95 $\pm$ 0.23          & \textbf{0.60 $\pm$ 0.50} (TP$_{13}$) & 0.16 $\pm$ 0.21$\ast$ & \textbf{0.30 $\pm$ 0.28} (TP$_{25}$) & 1.17 $\pm$ 1.69       \\
            \bottomrule
        \end{tabular}
        }
    \caption{Metrics M1-M3.}
    \end{subtable}
    \hfill
    \begin{subtable}[h]{\textwidth}
        \centering
        \resizebox{\textwidth}{!}{%
        \begin{tabular}{ccccccccccccc}
            \toprule
            \multicolumn{1}{c}{} &
              \multicolumn{2}{c}{\textbf{MSS}} &
              \multicolumn{2}{c}{\textbf{MMCR}} &
              \multicolumn{2}{c}{\textbf{MSp}} \\
             & \textbf{Best TP} & \textbf{TP$_5$} & \textbf{Best TP} & \textbf{TP$_5$} & \textbf{Best TP} & \textbf{TP$_5$} \\
            \midrule
            \multicolumn{1}{c}{$\widehat{\mathbf{S}}$} & \textbf{0.12 $\pm$ 0.10} (TP$_{5}$) & \textbf{0.12 $\pm$ 0.10} & \textbf{0.02 $\pm$ 0.02} (TP$_{4}$) & 0.91 $\pm$ 2.90 & \textbf{0.16 $\pm$ 0.12} (TP$_{4}$) & 0.09 $\pm$ 0.08 \\
            \midrule
            \textbf{S1}                     & \textbf{0.15 $\pm$ 0.10} (TP$_{5}$)  & \textbf{0.15 $\pm$ 0.10} & \textbf{0.02 $\pm$ 0.02} (TP$_{4}$)  & 0.42 $\pm$ 0.80       & \textbf{0.23 $\pm$ 0.12} (TP$_{4}$)  & 0.08 $\pm$ 0.06           \\
            \textbf{S2}                     & \textbf{0.14 $\pm$ 0.09} (TP$_{4}$)  & 0.06 $\pm$ 0.10$\ast$    & \textbf{0.02 $\pm$ 0.02} (TP$_{4}$)  & 3.81 $\pm$ 7.79$\ast$ & \textbf{0.12 $\pm$ 0.07} (TP$_{4}$)  & 0.04 $\pm$ 0.05$\ast$     \\
            \textbf{S3}                     & \textbf{0.07 $\pm$ 0.07} (TP$_{5}$)  & \textbf{0.07 $\pm$ 0.07} & \textbf{0.02 $\pm$ 0.01} (TP$_{4}$)  & 0.81 $\pm$ 2.23$\ast$ & \textbf{0.05 $\pm$ 0.04} (TP$_{5}$)  & \textbf{0.05 $\pm$ 0.04}  \\
            \textbf{S4}                     & \textbf{0.14 $\pm$ 0.07} (TP$_{9}$)  & 0.13 $\pm$ 0.11$\ast$    & \textbf{0.02 $\pm$ 0.01} (TP$_{4}$)  & 0.97 $\pm$ 1.87$\ast$ & \textbf{0.21 $\pm$ 0.10} (TP$_{9}$)  & 0.07 $\pm$ 0.08$\ast$     \\
            \textbf{S5}                     & \textbf{0.18 $\pm$ 0.06} (TP$_{1}$)  & 0.16 $\pm$ 0.03          & \textbf{0.01 $\pm$ 0.00} (TP$_{9}$)  & 0.01 $\pm$ 0.01       & \textbf{0.26 $\pm$ 0.08} (TP$_{9}$)  & 0.17 $\pm$ 0.07$\ast$     \\
            \textbf{S6}                     & \textbf{0.08 $\pm$ 0.04} (TP$_{4}$)  & 0.05 $\pm$ 0.07$\ast$    & \textbf{0.02 $\pm$ 0.01} (TP$_{4}$)  & 2.05 $\pm$ 4.19$\ast$ & \textbf{0.23 $\pm$ 0.14} (TP$_{4}$)  & 0.23 $\pm$ 0.14           \\
            \textbf{S7}                     & \textbf{0.32 $\pm$ 0.06} (TP$_{18}$) & 0.26 $\pm$ 0.06          & \textbf{0.01 $\pm$ 0.01} (TP$_{21}$) & 0.01 $\pm$ 0.01       & \textbf{0.23 $\pm$ 0.04} (TP$_{4}$)  & 0.19 $\pm$ 0.04           \\
            \textbf{S8}                     & \textbf{0.06 $\pm$ 0.12} (TP$_{24}$) & 0.01 $\pm$ 0.02$\ast$    & \textbf{0.02 $\pm$ 0.01} (TP$_{4}$)  & 1.81 $\pm$ 0.77$\ast$ & \textbf{0.04 $\pm$ 0.02} (TP$_{24}$) & 0.04 $\pm$ 0.04$\ast$     \\
            \textbf{S9}                     & \textbf{0.11 $\pm$ 0.06} (TP$_{5}$)  & \textbf{0.11 $\pm$ 0.06} & \textbf{0.03 $\pm$ 0.01} (TP$_{4}$)  & 0.18 $\pm$ 0.54$\ast$ & \textbf{0.05 $\pm$ 0.02} (TP$_{5}$)  & \textbf{0.05 $\pm$ 0.02}  \\
            \textbf{S10}                    & \textbf{0.15 $\pm$ 0.04} (TP$_{19}$) & 0.06 $\pm$ 0.08$\ast$    & \textbf{0.01 $\pm$ 0.01} (TP$_{24}$) & 0.66 $\pm$ 0.60$\ast$ & \textbf{0.24 $\pm$ 0.10} (TP$_{24}$) & 0.06 $\pm$ 0.06$\ast$     \\
            \textbf{S11}                    & \textbf{0.18 $\pm$ 0.07} (TP$_{5}$)  & \textbf{0.18 $\pm$ 0.07} & \textbf{0.01 $\pm$ 0.01} (TP$_{25}$) & 0.06 $\pm$ 0.19       & \textbf{0.21 $\pm$ 0.10} (TP$_{24}$) & 0.15 $\pm$ 0.08           \\
            \bottomrule
        \end{tabular}
        }
        \caption{Metrics M4-M6.}
    \end{subtable}
    \caption{Statistical test results for baseline (highest mean) and TP$_5$ as presented in \cite{Strombom:2014}. Boldface text indicates the best performance, and an asterisk indicates a statistical significance difference.}
    \label{table:stats}
\end{table}

\begin{figure}
     \centering
     \begin{subfigure}[h]{0.475\textwidth}
         \centering
         \includegraphics[width=\textwidth]{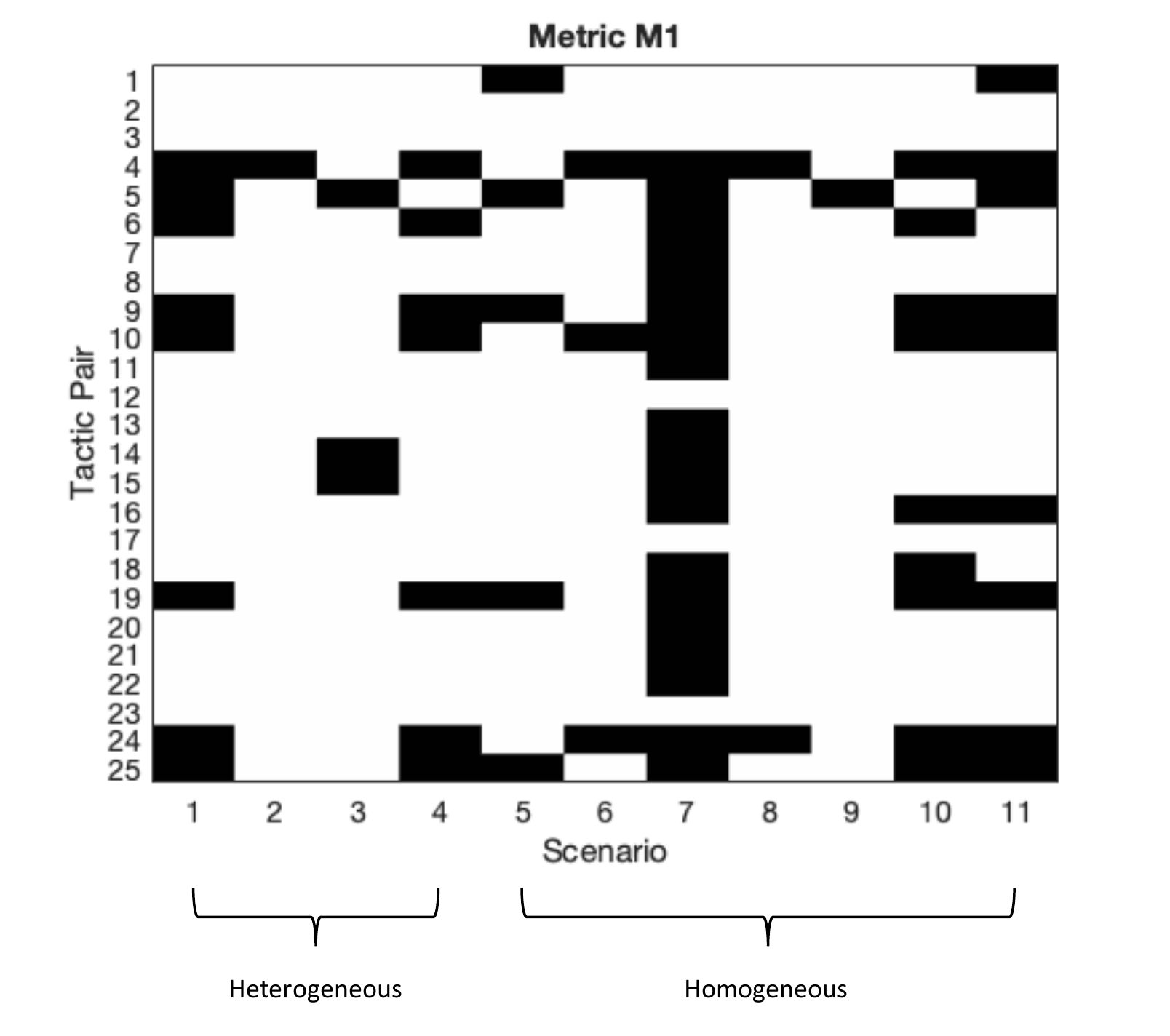}
         \caption{Mission Success for all scenarios.}
         \label{figure:M1}
     \end{subfigure}
     \hfill
     \begin{subfigure}[h]{0.475\textwidth}
         \centering
         \includegraphics[width=\textwidth]{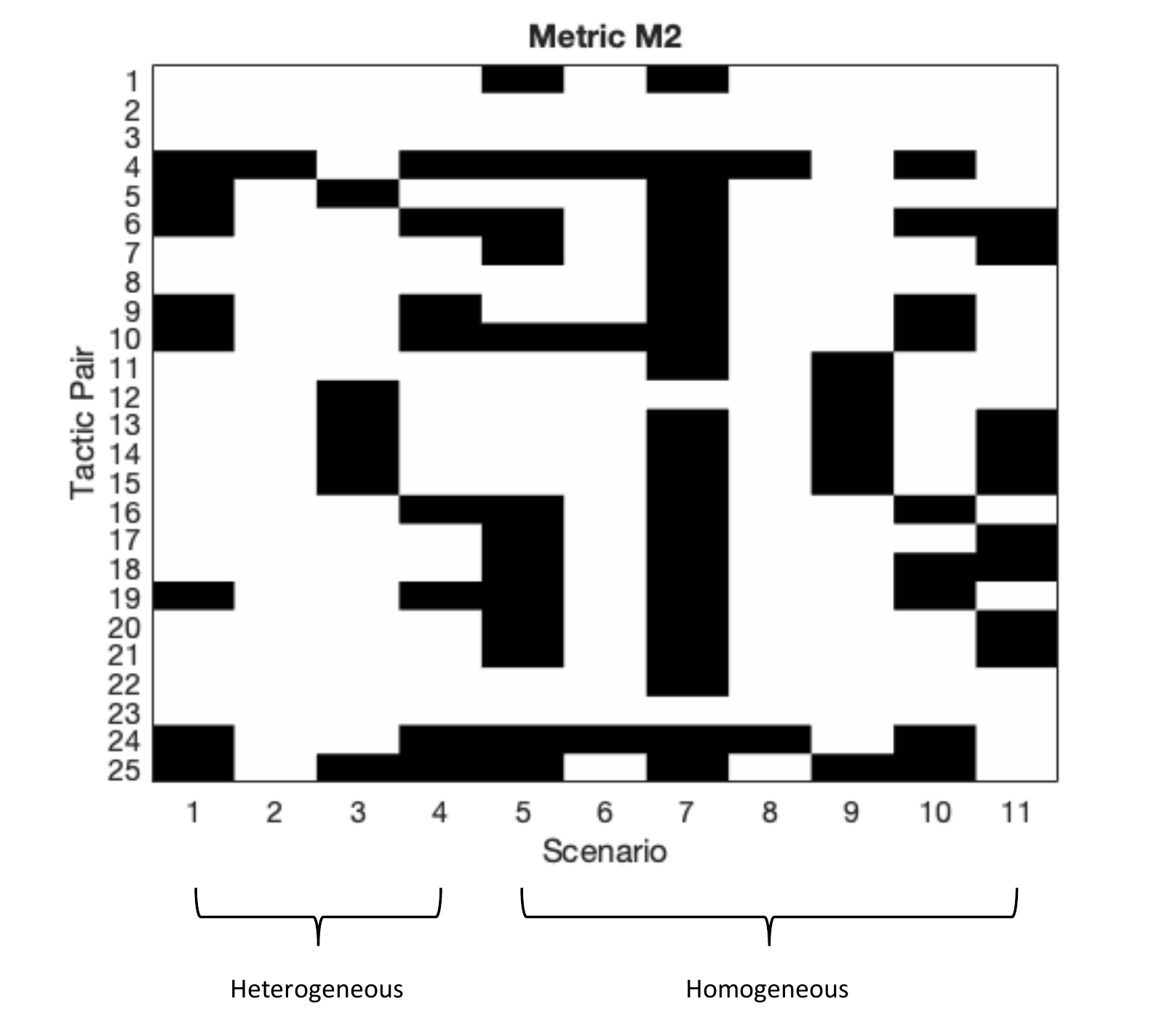}
         \caption{Mission Decision Stability for all scenarios.}
         \label{figure:M2}
     \end{subfigure}
     \hfill
     \begin{subfigure}[h]{0.475\textwidth}
         \centering
         \includegraphics[width=\textwidth]{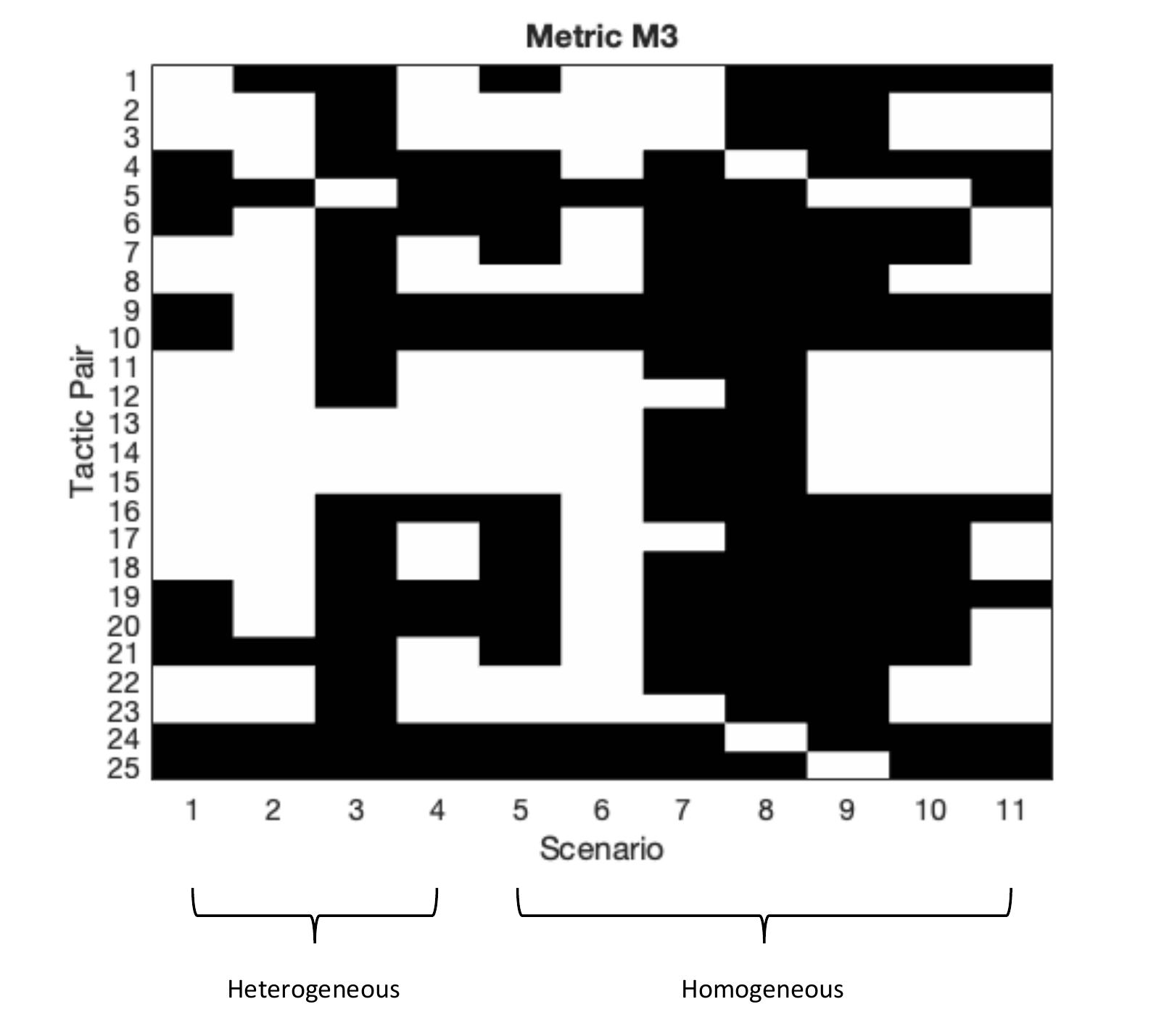}
         \caption{Swarm Decision Stability for all scenarios.}
         \label{figure:M3}
     \end{subfigure}
     \hfill
     \begin{subfigure}[h]{0.475\textwidth}
         \centering
         \includegraphics[width=\textwidth]{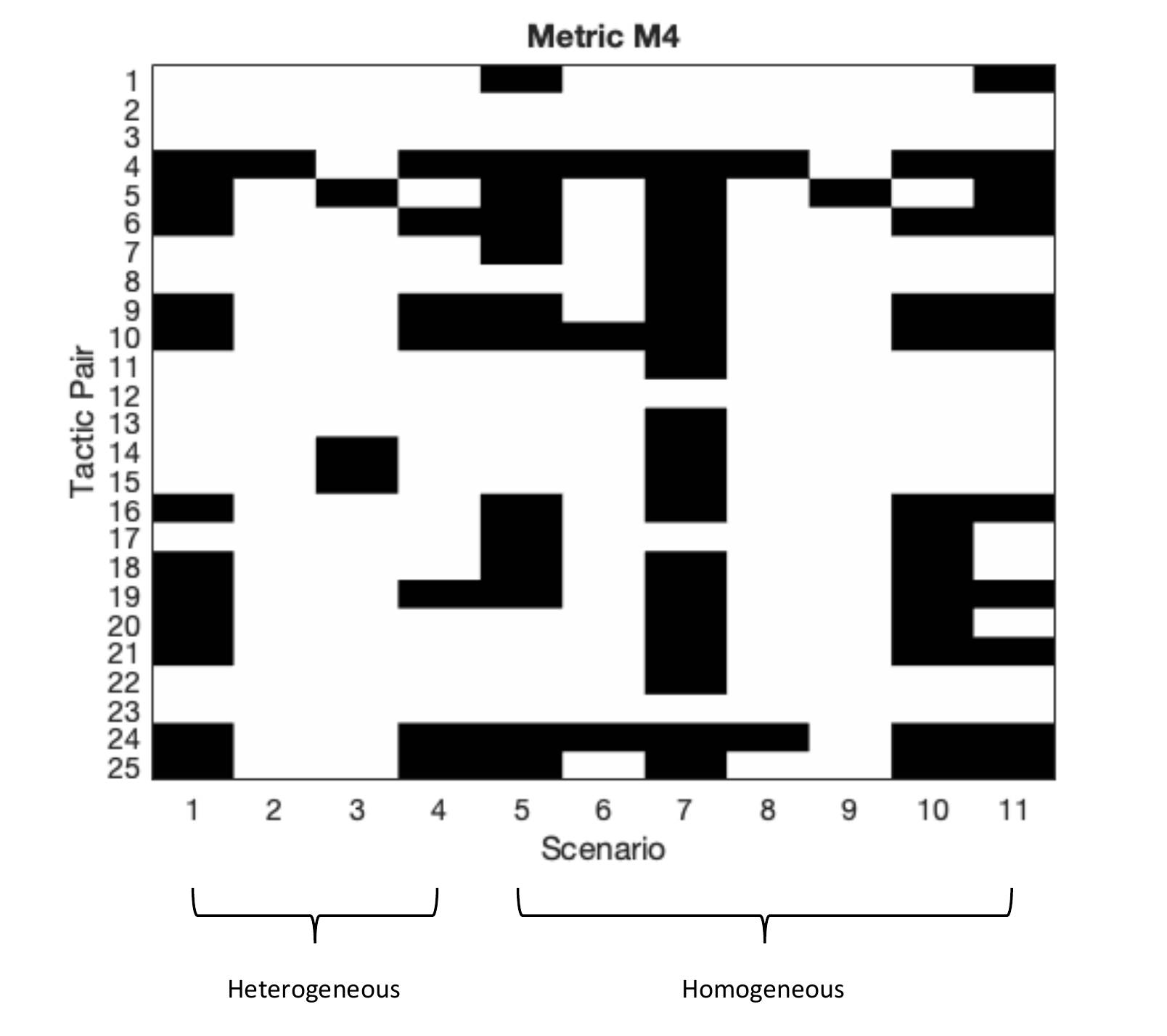}
         \caption{Mission Swarm Stability for all scenarios.}
         \label{figure:M4}
     \end{subfigure}
     \hfill
     \begin{subfigure}[h]{0.475\textwidth}
         \centering
         \includegraphics[width=\textwidth]{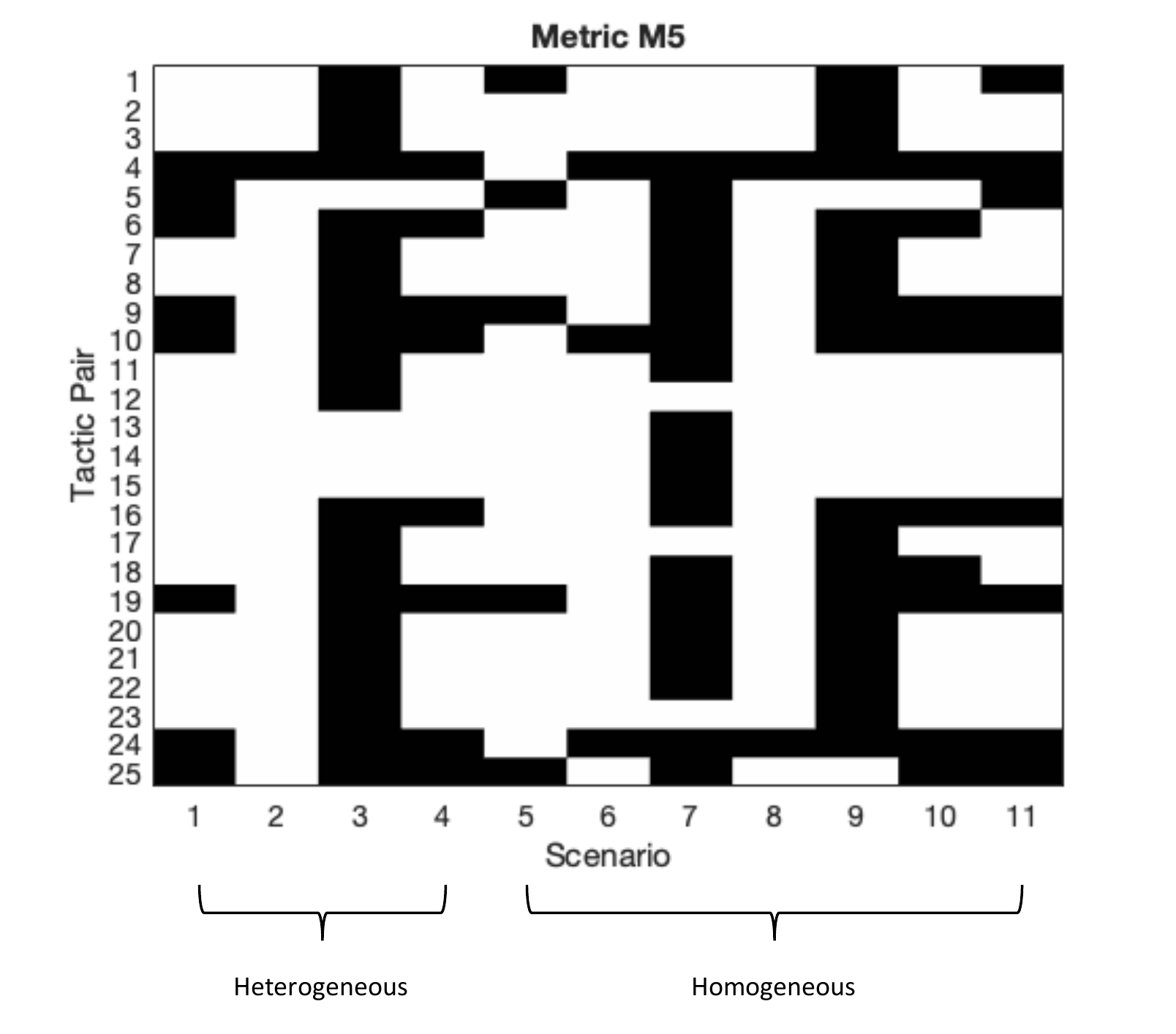}
         \caption{Mission Completion Rate for all scenarios.}
         \label{figure:M5}
     \end{subfigure}
     \hfill
     \begin{subfigure}[h]{0.475\textwidth}
         \centering
         \includegraphics[width=\textwidth]{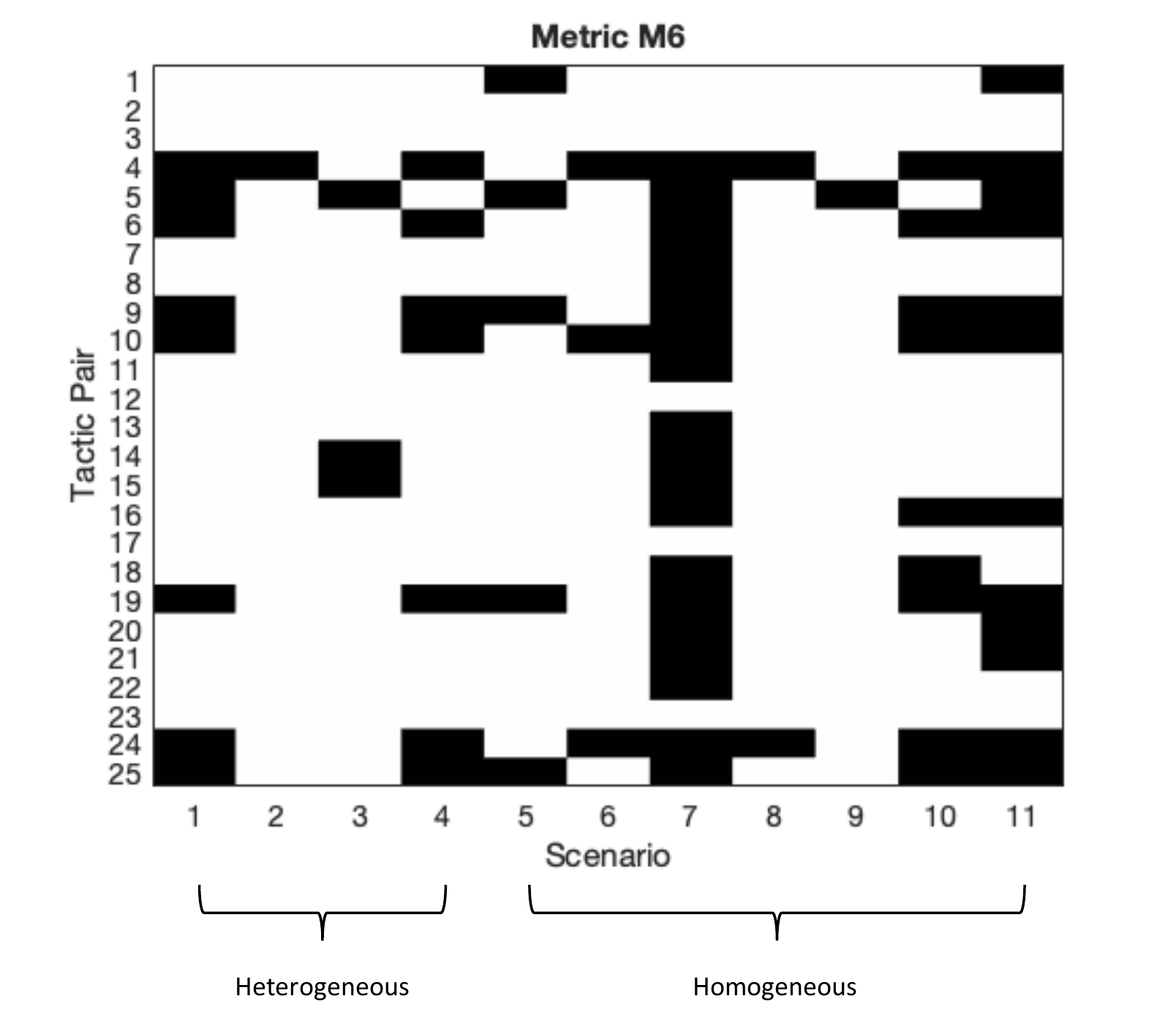}
         \caption{Mission Speed Ratio for all scenarios.}
         \label{figure:M6}
     \end{subfigure}
     \caption{Columns represent scenarios with rows representing each TP. This provides us with intuition to compare a TP across scenarios, or how a particular set of TPs performs for a specific scenario. The white segments represent that a TP is unsuitable for the scenario (statistical significance), with black representing suitability for control.}
     \label{fig:BHOT}
\end{figure}

We have focused on TP$_5$ in Table~\ref{table:stats} due to its significance across the literature and use throughout several studies. Figure~\ref{fig:BHOT} represents the aggregation and replication of this statistical analysis across all metrics and for all TPs. To read each sub-figure, white segments represent that a TP is not suitable for the scenario (i.e. statistical significance worse), with black segments highlighting that a TP is suitable for control. In two ways, we summarise the sub-figures of Figure~\ref{fig:BHOT}. The first is to consider a TPs performance over multiple scenarios and metrics. The second is considering the set of TPs control-suitable for a specific scenario. This offers insight into selecting TPs for particular contexts where an exact scenario cannot be determined. In this case, it is possible to determine a default TP to control the swarm for heterogeneous, homogeneous or an unknown setting.

For instance, if we consider Figure~\ref{figure:M1} from a column-wise (scenario) perspective, we observe that a limited subset of TPs is suitable for control across each scenario. However, for the case of S7, this does not hold, as most TPs are suitable for control. From a row-wise (TP) perspective, we observe that TPs are often suitable for control in most scenarios (for example, TP$_4$ and TP$_5$) or unsuitable in most scenarios (for example, TP$_{12}$ and TP$_{13}$). Considering each metric represented in Figure~\ref{fig:BHOT} through the lens of swarm control, we can now develop behavioural libraries for using each TP for particular scenarios.

In this section, we have systematically explored the parameterisation of autonomous behaviours (atomic collect and drive actions) for a shepherding agent to control a swarm in both homogeneous and heterogeneous settings. We have uncovered that a rule-based approach with single-instance implementations of autonomous atomic behaviours is not optimal when classic shepherding context assumptions are relaxed; the swarm becomes heterogeneous. However, through systematic variations of behaviour combinations, we have demonstrated that mission completion remains high when variations of autonomous behaviours are used, an invariant of the metric to assess performance. The upshot is that a slight variation in the parameterisation of the atomic collect or drive behaviour is sufficient to minimise performance degradation across these scenarios. This quantifies that it can effectively and efficiently control swarms with disparate agent-type distributions in shepherding contexts. The remaining challenge for a control agent is to identify the swarm's situation so that it can deliver meaningful real-time guidance from recognition~\cite{Bakar2016}. In the following section, we present an intelligent control agent. The intelligent control agent recognises the swarm context and uses this information to select an appropriate TP response.

\section{AI2A: Actionable-Information to Actuators}\label{sec:AI2A}

In this section, we quantify the effect of our context-awareness system on swarm control, augmenting a classic reactive agent with cognitive capabilities. We consider three lenses with which to measure this effect, mission effectiveness, control stability, and herding efficiency. These lenses extend the metrics we present in Table~\ref{table:metrics}, including two additional aspects we discuss below.

Mission effectiveness informs on the performance of the context-aware agent to achieve $P_\Pi - P_G < R_G$, quantifying this relative to a classic control agent without a context-aware system. Control stability measures the impact of the context-aware control agent\textquoteright s tactic-pair selection and behaviour modulation on the swarm and mission. Our final lens, herding efficiency, explores the impact of the context-aware system on physical aspects of the system, such as distance travelled and speed of movement. Each of our lenses explores 11 scenarios (4 heterogeneous, S1-S4, and 7 homogeneous, S5-S11) for evaluation, comparing a reactive control agent with and without augmentation of our context-awareness system. 110 trials were completed for each setting (with and without context awareness), yielding 220 trials across 11 scenarios with swarm size, $N=20$. Each scenario-pair setting (with and without context awareness) is parameterised identically, such that both settings start with a classic reactive control agent ($\sigma_1^{D100}, \sigma_2^{F2G}, \sigma^{c_2} = 1$ and $\sigma^{c_3} = 1$) per~\cite{Strombom:2014}. For trials with the context-awareness system present, the agent is parameterised after the initial observation period of $\sigma^{c_1}$. Table~\ref{table:evaluation} summarises our experimentation and analysis objectives in this study.

\begin{table}[h!]
    \centering
    \def\arraystretch{1}
    \resizebox{\textwidth}{!}{%
    \begin{tabular}{cL{20em}L{20em}}
        \toprule
        No. & Hypothesis & Metrics \\ \midrule
        1 & A Context-aware system will increase the shepherding agent mission performance & MS and Mission Length (\cite{8228268, Himo2022:HeterogeneousResponse, Strombom:2014}), and Number of $\pi_i$ influenced by $\beta$ (\cite{Strombom:2014})\\
        2 & A context-aware system will increase the shepherding agent control stability for herding tasks & number of separated swarm agents (\cite{Goel:2019}), MDS, DSS, and MSS \\
        3 & A context-aware system will increase the shepherding agent herding efficiency & $\Pi$ and $\beta$ distance travelled (\cite{8228268}), MSP, and MCR \\
        \bottomrule
    \end{tabular}
    }
    \caption{Summary of experiment design and evaluation measures to quantify the effect of a context-awareness system for swarm control.}
    \label{table:evaluation}
\end{table}

\subsection{Mission Effectiveness}

Our mission effectiveness analysis considers four aspects; these results are depicted in Table~\ref{table:effectiveness}. The first is mission success rate, reported as the percentage of trials where $\vert\vert P_\Pi - P_G \vert\vert < R_G$ given $t \leq T_{\text{max}}$. The second is the mission length across all trials (successes and failures); as with previous studies, this is the number of simulation steps. The third measure is mission length across the subset of trials for the cases where mission success was achieved. The fourth and final measure of mission effectiveness is the number of $\pi_i$ influenced by $\beta$; \cite{Goel:2019} note that \enquote{A precise measure of influence using leaders or predators or a combination of leaders and predators to achieve the mission is not adequately studied.}~(pg.1). We use the interaction radius of $\beta$ to determine if a $\pi_i$ is directly influenced, or not; the influence range parameterisation is consistent with that described in~\cite{Strombom:2014}.

We observe an overall improvement in mission success of 10\% for the control agent with the context-aware system; most notably, the best mission success is achieved with heterogeneous swarm scenarios (S1-S4), outperforming the agent without the context-awareness system for all heterogeneous settings. In addition, the context-aware agent observes a decrease in mission length for both all trials and only successful trials, with statistically significant improvement in 72.7\% of all trials and 81.8\% of successful trials. Finally, for our final mission effectiveness measure (number of $\pi$ influenced by $\beta$), we observe statistical significance in 100\% of trials, indicating that the context-aware control agent influences a more significant number of swarm agents across all homogeneous and heterogeneous scenarios.

\begin{table}[h!]
    \centering
    \def\arraystretch{1.25}
    \resizebox{\textwidth}{!}{%
    \begin{tabular}{lcccccccc}
        \toprule
        \multicolumn{9}{c}{\textbf{Metrics}} \\
        \multicolumn{1}{c}{} &
          \multicolumn{2}{c}{\textbf{Mission Success Rate}} &
          \multicolumn{2}{c}{\textbf{Mission Length (All Trials)}} &
          \multicolumn{2}{c}{\textbf{Mission Length (Successful Trials)}} &
          \multicolumn{2}{c}{\textbf{Num. $\pi_i$ Influenced by $\beta$}} \\ &
          \multicolumn{1}{c}{\textbf{With}} &
          \multicolumn{1}{c}{\textbf{Without}} &
          \multicolumn{1}{c}{\textbf{With}} &
          \multicolumn{1}{c}{\textbf{Without}} &
          \multicolumn{1}{c}{\textbf{With}} &
          \multicolumn{1}{c}{\textbf{Without}} &
          \multicolumn{1}{c}{\textbf{With}} &
          \multicolumn{1}{c}{\textbf{Without}} \\
        \midrule
        $\widehat{\mathbf{S}}$  & \textbf{74\% $\pm$ 28\%} & 64\% $\pm$ 33\% & \textbf{2,157.50 $\pm$ 1,959.90}  & 3,320.10 $\pm$ 1,646.60$\ast$ & \textbf{1,086.50 $\pm$ 909.98}   & 2,275.10 $\pm$ 1,111.80$\ast$ & \textbf{7.17 $\pm$ 8.52} & 3.9 $\pm$ 6.37$\ast$ \\
        \midrule
        \textbf{S1}  & \textbf{90\%}  & 70\%           & \textbf{1,136.40 $\pm$ 1,414.50}     & 2,956.30 $\pm$ 1,635.00$\ast$ & \textbf{690.60 $\pm$   121.40}   & 2,016.60 $\pm$   758.80$\ast$ & \textbf{13.05 $\pm$ 8.61} & 3.84 $\pm$ 5.44$\ast$ \\
        \textbf{S2}  & \textbf{100\%} & 30\%           & \textbf{665.00 $\pm$ 0138.00}        & 4,721.40 $\pm$ 852.60$\ast$   & \textbf{665.00 $\pm$   138.00}   & 3,723.70 $\pm$ 1,066.80$\ast$ & \textbf{11.31 $\pm$ 8.41} & 2.99 $\pm$ 6.19$\ast$ \\
        \textbf{S3}  & \textbf{100\%} & 90\%           & \textbf{863.10 $\pm$ 0271.90}        & 2,229.00 $\pm$ 1,077.50$\ast$ & \textbf{863.10 $\pm$   271.90}   & 1,904.60 $\pm$   349.00$\ast$ & \textbf{8.87 $\pm$ 8.27}  & 4.22 $\pm$ 6.66$\ast$ \\
        \textbf{S4}  & \textbf{50\%}  & 30\%           & \textbf{3,702.60 $\pm$ 1,674.30}     & 4,393.40 $\pm$ 1,496.90       & \textbf{2,256.20 $\pm$ 1,037.70} & 2,630.30 $\pm$ 1,850.10       & \textbf{5.43 $\pm$  7.9}  & 3.32 $\pm$ 6.24$\ast$ \\
        \textbf{S5}  & 40\%           & \textbf{50\%}  & \textbf{4,099.00 $\pm$ 1,391.90}     & 4,507.90 $\pm$ 741.20         & \textbf{2,524.00 $\pm$   547.20} & 3,866.80 $\pm$   456.80$\ast$ & \textbf{3.79 $\pm$ 6.67}  & 3.48 $\pm$ 5.33$\ast$ \\
        \textbf{S6}  & 80\%           & \textbf{90\%}  & \textbf{1,617.60 $\pm$ 1,866.70}     & 2,571.30 $\pm$ 1,091.30$\ast$ & \textbf{734.80 $\pm$   161.60}   & 2,284.90 $\pm$   645.70$\ast$ & \textbf{9.41 $\pm$ 9.32}  & 4.08 $\pm$  6.6$\ast$ \\
        \textbf{S7}  & 90\%           & \textbf{100\%} & \textbf{989.00 $\pm$ 1,462.60}       & 1,891.60 $\pm$ 1,022.30$\ast$ & \textbf{526.80 $\pm$    53.90}   & 1,891.60 $\pm$ 1,022.30$\ast$ & \textbf{12.87 $\pm$ 7.31} & 6.21 $\pm$ 7.94$\ast$ \\
        \textbf{S8}  & \textbf{90\%}  & 80\%           & \textbf{1,187.50 $\pm$ 1,407.10}     & 2,447.20 $\pm$ 1,490.40$\ast$ & \textbf{747.30 $\pm$   218.80}   & 1,771.70 $\pm$   499.00$\ast$ & \textbf{11.65 $\pm$ 9.05} & 4.35 $\pm$ 5.91$\ast$ \\
        \textbf{S9}  & 100\%          & 100\%          & \textbf{768.70 $\pm$ 73.60}          & 1,182.80 $\pm$ 202.70$\ast$   & \textbf{768.70 $\pm$    73.60}   & 1,182.80 $\pm$   202.70$\ast$ & \textbf{14.15 $\pm$ 8.05} & 9.17 $\pm$ 9.06$\ast$ \\
        \textbf{S10} & \textbf{50\%}  & 0\%            & \textbf{3,879.20 $\pm$ 1,696.20}     & 5,149.00 $\pm$      0$\ast$   & \textbf{2,609.40 $\pm$ 1,562.90} &        N/A$\ast$              & \textbf{8.8 $\pm$ 8.97}   & 3.88 $\pm$ 6.79$\ast$ \\
        \textbf{S11} & 20\%           & \textbf{60\%}  & 4,824.90 $\pm$ 796.30                & \textbf{4,471.6 $\pm$ 696.30} & \textbf{3,528.50 $\pm$ 1,226.80} & 4,020.00 $\pm$   510.80       & \textbf{3.66 $\pm$ 6.14}  & 3.02 $\pm$    5$\ast$ \\
        \bottomrule
    \end{tabular}
    }
    \caption{Mission effectiveness analysis of a swarm control agent with and without the presented context-awareness system. A star indicates that the difference is statistically significant.}
    \label{table:effectiveness}
\end{table}

\subsection{Control Stability}

We consider four metrics to analyse control stability, including three metrics as presented in Table~\ref{table:metrics}; these results are depicted in Table~\ref{table:stability}. The first metric is the mean number of separated $\pi_i$ agents from the central swarm cluster $\Pi$ over each trial. We observe an overall lower mean number of separated agents for the control agent with context awareness. Mission decision stability effectively measures the sensitivity of the mission outcome to the number of behaviour changes, manifesting as the count of $\sigma_1 \rightarrow \sigma_2$ and $\sigma_2 \rightarrow \sigma_1$ changes. We observe significance in favour of the control agent with the context-awareness system in 45\% of scenarios, achieving the best outcome in 64\% of all scenarios.

Decision swarm stability is the ratio between the number of separated $\pi_i$ and the number of decision changes $\beta$ conducts, averaged over the total mission. The control agent without context awareness performs marginally better in 55\% of scenarios, although the significance is only recorded in 27\%. Of interest, the control agent with context awareness has better aggregate performance across all scenarios; however, the results are not significant. The final metric is mission swarm stability, measuring the sensitivity of mission success to the number of separated swarm agents. As with decision swarm stability, we observe a marginal distinction between the control agent with and without context awareness. In this case, the context-aware enabled agent is best in 55\% of scenarios, with significance recorded in 18\%.

\begin{table}[h!]
    \centering
    \def\arraystretch{1.25}
    \resizebox{\textwidth}{!}{%
    \begin{tabular}{lcccccccc}
        \toprule
        \multicolumn{9}{c}{\textbf{Metrics}} \\
        \multicolumn{1}{c}{} &
          \multicolumn{2}{c}{\textbf{Mean. Sep Agents}} &
          \multicolumn{2}{c}{\textbf{MDS}} &
          \multicolumn{2}{c}{\textbf{DSS}} &
          \multicolumn{2}{c}{\textbf{MSS}} \\
        \multicolumn{1}{c}{} &
          \multicolumn{1}{c}{\textbf{With}} &
          \multicolumn{1}{c}{\textbf{Without}} &
          \multicolumn{1}{c}{\textbf{With}} &
          \multicolumn{1}{c}{\textbf{Without}} &
          \multicolumn{1}{c}{\textbf{With}} &
          \multicolumn{1}{c}{\textbf{Without}} &
          \multicolumn{1}{c}{\textbf{With}} &
          \multicolumn{1}{c}{\textbf{Without}} \\
        \midrule
        $\widehat{\mathbf{S}}$  & \textbf{7.18 $\pm$ 2.21} & 7.70 $\pm$ 2.26 & \textbf{0.29 $\pm$ 0.29} & 0.24 $\pm$ 0.24 & \textbf{1.66 $\pm$ 1.30} & 2.30 $\pm$ 2.48 & \textbf{0.10 $\pm$ 0.05} & 0.09 $\pm$ 0.06 \\
        \midrule
        \textbf{S1}     & 8.58 $\pm$ 2.59$\ast$    & \textbf{6.73 $\pm$ 2.40} & \textbf{0.38 $\pm$ 0.43}  & 0.15 $\pm$ 0.32$\ast$    & 2.12 $\pm$ 1.86          & \textbf{1.69 $\pm$ 3.00} & \textbf{0.10 $\pm$ 0.03} & 0.10 $\pm$ 0.08          \\
        \textbf{S2}     & 9.49 $\pm$ 0.76$\ast$    & \textbf{4.81 $\pm$ 1.15} & \textbf{0.14 $\pm$ 0.08}  & 0.01 $\pm$ 0.02$\ast$    & 1.13 $\pm$ 0.57$\ast$    & \textbf{0.18 $\pm$ 0.11} & 0.11 $\pm$ 0.01          & \textbf{0.06 $\pm$ 0.10} \\
        \textbf{S3}     & 7.32 $\pm$ 1.76          & \textbf{5.89 $\pm$ 2.20} & \textbf{0.55 $\pm$ 0.39}  & 0.26 $\pm$ 0.40$\ast$    & 2.37 $\pm$ 1.42$\ast$    & \textbf{1.04 $\pm$ 1.26} & \textbf{0.14 $\pm$ 0.03} & 0.15 $\pm$ 0.07          \\
        \textbf{S4}     & \textbf{4.56 $\pm$ 2.56} & 8.43 $\pm$ 2.60$\ast$    & 0.04 $\pm$ 0.08           & \textbf{0.21 $\pm$ 0.37} & \textbf{0.34 $\pm$ 0.41} & 4.30 $\pm$ 4.76$\ast$    & 0.08 $\pm$ 0.09          & \textbf{0.03 $\pm$ 0.05} \\
        \textbf{S5}     & \textbf{6.01 $\pm$ 2.04} & 11.97 $\pm$ 3.33$\ast$   & 0.09 $\pm$ 0.12           & \textbf{0.48 $\pm$ 0.49} & \textbf{1.15 $\pm$ 1.08} & 7.16 $\pm$ 6.02$\ast$    & 0.05 $\pm$ 0.07          & \textbf{0.05 $\pm$ 0.05} \\
        \textbf{S6}     & 7.63 $\pm$ 3.36          & \textbf{6.63 $\pm$ 2.47} & \textbf{0.23 $\pm$ 0.29}  & 0.19 $\pm$ 0.31          & 1.51 $\pm$ 1.38          & \textbf{1.16 $\pm$ 1.91} & 0.09 $\pm$ 0.05$\ast$    & \textbf{0.14 $\pm$ 0.06} \\
        \textbf{S7}     & 8.63 $\pm$ 2.75$\ast$    & \textbf{7.18 $\pm$ 1.66} & \textbf{0.23 $\pm$ 0.28}  & 0.09 $\pm$ 0.07          & 1.60 $\pm$ 1.42$\ast$    & \textbf{0.50 $\pm$ 0.33} & 0.10 $\pm$ 0.04$\ast$    & \textbf{0.15 $\pm$ 0.06} \\
        \textbf{S8}     & 9.60 $\pm$ 3.05          & \textbf{8.18 $\pm$ 2.86} & \textbf{0.51 $\pm$ 0.43}  & 0.08 $\pm$ 0.10$\ast$    & 3.11 $\pm$ 2.12$\ast$    & \textbf{0.74 $\pm$ 0.92} & \textbf{0.09 $\pm$ 0.03} & 0.09 $\pm$ 0.05          \\
        \textbf{S9}     & 5.69 $\pm$ 0.82          & \textbf{5.20 $\pm$ 0.96} & 0.65 $\pm$ 0.38           & \textbf{0.92 $\pm$ 0.25} & \textbf{2.07 $\pm$ 0.88} & 2.45 $\pm$ 0.75          & \textbf{0.18 $\pm$ 0.03} & 0.20 $\pm$ 0.04          \\
        \textbf{S10}    & \textbf{5.88 $\pm$ 2.37} & 9.06 $\pm$ 2.99$\ast$    & \textbf{0.29 $\pm$ 0.40}  & 0 $\pm$    0$\ast$       & \textbf{1.78 $\pm$ 1.64} & 2.11 $\pm$ 3.42          & \textbf{0.08 $\pm$ 0.09} & 0 $\pm$    0$\ast$       \\
        \textbf{S11}    & \textbf{5.53 $\pm$ 2.28} & 10.59 $\pm$ 2.26$\ast$   & 0.10 $\pm$ 0.32$\ast$     & \textbf{0.22 $\pm$ 0.33} & \textbf{1.04 $\pm$ 1.52} & 3.96 $\pm$ 4.77$\ast$    & \textbf{0.03 $\pm$ 0.08} & 0.06 $\pm$ 0.06          \\
        \bottomrule
    \end{tabular}
    }
    \caption{Stability analysis of a swarm control agent with and without the presented context-awareness system.}
    \label{table:stability}
\end{table}

\subsection{Herding Efficiency}

Our final perspective is herding efficiency, which captures particular physically-interpretable aspects of the system; results are depicted in Table~\ref{table:efficiency}. The first metric is the total swarm distance moved, measuring the cumulative path distance from $P^{t=0}_\Pi$ to $P^{t=T}_\Pi$. We calculate $P^t_\Pi = \sfrac{1}{N}\sum_{i=1}^N{P^t_{\pi_i}}$. We observe statistical significance across all scenarios, with the control agent without the context-aware system minimising the total distance travelled by the swarm. Comparing the settings with and without context awareness, we note that the context-aware system results have a significantly lower deviation in the distance travelled per situation scenario ($Z = 2.63, p < 0.01$). This is an important finding in our work concerning system energy utilisation. For instance, in a physical robotic system, it is crucial to ensure that energy use is predictable, enabling performance guarantees to be placed on the system. While a control agent with our context-awareness system increases the distance the swarm travels, the empirical variance is stable between scenarios. The control agent without context awareness observes a standard deviation approximately the magnitude of the mean distance travelled (see S5 for example), with a lower bound approaching that observed for the control agent with context awareness. The second factor of further consideration here is regarding the control agent type implementation, particularly that for the control agent without context awareness as defined conditions for $\sigma_2$ must be met prior to the execution of $\sigma_1$.

Our second metric is the control agent\textquoteright s total distance travelled, calculated as per the swarm total distance travelled; however, utilising only $P_{\beta}$. For this metric, we see the statistical significance in only one scenario; for all other settings, the total distance each agent type moves is approximately equivalent. Our third metric is mission speed, in which we observe significance in over 90\% of scenarios, with the context-aware enabled agent demonstrating higher speed over different scenarios. Our final metric is the mission completion rate. This metric provides insight into the control strategy efficiency, mainly when the mission is unsuccessful. The control agent with a context-awareness system observes the best performance in 90\% of scenarios, although the significance is recorded in only 36\%.

\begin{table}[h!]
    \centering
    \def\arraystretch{1.25}
    \resizebox{\textwidth}{!}{%
    \begin{tabular}{lcccccccc}
        \toprule
        \multicolumn{9}{c}{\textbf{Metrics}} \\
        \multicolumn{1}{c}{} &
          \multicolumn{2}{c}{\textbf{Swarm Total Dist.}} &
          \multicolumn{2}{c}{\textbf{Agent Total Dist.}} &
          \multicolumn{2}{c}{\textbf{MSp}} &
          \multicolumn{2}{c}{\textbf{MCR}} \\
        \multicolumn{1}{c}{} &
          \multicolumn{1}{c}{\textbf{With}} &
          \multicolumn{1}{c}{\textbf{Without}} &
          \multicolumn{1}{c}{\textbf{With}} &
          \multicolumn{1}{c}{\textbf{Without}} &
          \multicolumn{1}{c}{\textbf{With}} &
          \multicolumn{1}{c}{\textbf{Without}} &
          \multicolumn{1}{c}{\textbf{With}} &
          \multicolumn{1}{c}{\textbf{Without}} \\
        \midrule
        $\widehat{\mathbf{S}}$   & 171.60 $\pm$ 11.20$\ast$ & \textbf{99.58 $\pm$ 34.70} & 226.52 $\pm$ 29.41 & 227.24 $\pm$ 62.60 & \textbf{0.18 $\pm$ 0.60} & 0.07 $\pm$ 0.03$\ast$ & \textbf{0.03 $\pm$ 0.03} & 0.97 $\pm$ 1.26$\ast$ \\
        \midrule
        \textbf{S1}     & 175.39 $\pm$ 10.67$\ast$ & \textbf{97.38 $\pm$ 37.11}  & \textbf{235.13 $\pm$ 16.63} & 248.03 $\pm$ 38.20          & \textbf{0.24 $\pm$ 0.09} & 0.07 $\pm$ 0.04$\ast$ & \textbf{0.01 $\pm$ 0.01} & 0.48 $\pm$ 0.81       \\
        \textbf{S2}     & 173.59 $\pm$ 10.07$\ast$ & \textbf{133.07 $\pm$ 21.56} & \textbf{227.23 $\pm$ 10.69} & 248.59 $\pm$ 83.37$\ast$    & \textbf{0.27 $\pm$ 0.04} & 0.03 $\pm$ 0.01$\ast$ & \textbf{0.01 $\pm$    0} & 1.08 $\pm$ 0.96$\ast$ \\
        \textbf{S3}     & 173.59 $\pm$ 10.08$\ast$ & \textbf{88.30 $\pm$ 25.70}  & \textbf{225.32 $\pm$ 15.81} & 239.09 $\pm$ 35.45          & \textbf{0.23 $\pm$ 0.09} & 0.09 $\pm$ 0.03$\ast$ & \textbf{0.01 $\pm$ 0.01} & 0.07 $\pm$ 0.19       \\
        \textbf{S4}     & 168.82 $\pm$ 15.45$\ast$ & \textbf{67.33 $\pm$ 53.43}  & 211.89 $\pm$ 36.72          & \textbf{196.19 $\pm$ 77.75} & \textbf{0.06 $\pm$ 0.05} & 0.04 $\pm$ 0.05$\ast$ & \textbf{0.04 $\pm$ 0.06} & 3.45 $\pm$ 4.39$\ast$ \\
        \textbf{S5}     & 159.24 $\pm$ 11.10$\ast$ & \textbf{56.20 $\pm$ 51.69}  & 235.49 $\pm$ 37.04          & \textbf{216.65 $\pm$ 88.45} & \textbf{0.05 $\pm$ 0.02} & 0.03 $\pm$ 0.02       & \textbf{0.07 $\pm$ 0.07} & 2.08 $\pm$ 3.48       \\
        \textbf{S6}     & 172.37 $\pm$ 10.12$\ast$ & \textbf{103.96 $\pm$ 53.97} & \textbf{226.91 $\pm$ 31.10} & 247.91 $\pm$ 34.92          & \textbf{0.21 $\pm$ 0.11} & 0.08 $\pm$ 0.03$\ast$ & \textbf{0.02 $\pm$ 0.01} & 0.05 $\pm$ 0.13$\ast$ \\
        \textbf{S7}     & 175.44 $\pm$ 10.39$\ast$ & \textbf{141.17 $\pm$  9.35} & 222.63 $\pm$ 22.57          & \textbf{212.48 $\pm$ 74.71} & \textbf{0.31 $\pm$ 0.10} & 0.11 $\pm$ 0.04$\ast$ & 0.02 $\pm$ 0.01$\ast$    & \textbf{0.01 $\pm$ 0} \\
        \textbf{S8}     & 174.80 $\pm$ 11.13$\ast$ & \textbf{125.69 $\pm$ 25.37} & 235.28 $\pm$ 15.14          & \textbf{208.97 $\pm$ 74.12} & \textbf{0.23 $\pm$ 0.09} & 0.09 $\pm$ 0.05$\ast$ & \textbf{0.02 $\pm$ 0.01} & 0.28 $\pm$ 0.61       \\
        \textbf{S9}     & 176.36 $\pm$ 10.05$\ast$ & \textbf{60.02 $\pm$  9.89}  & \textbf{221.15 $\pm$  7.50} & 225.30 $\pm$  6.58          & \textbf{0.23 $\pm$ 0.01} & 0.15 $\pm$ 0.02$\ast$ & \textbf{0.01 $\pm$ 0.01} & 0.02 $\pm$ 0.01       \\
        \textbf{S10}    & 173.04 $\pm$  9.26$\ast$ & \textbf{110.95 $\pm$ 49.38} & 221.67 $\pm$ 37.74          & \textbf{190.38 $\pm$ 07.93} & \textbf{0.06 $\pm$ 0.04} & 0.02 $\pm$ 0.01$\ast$ & \textbf{0.04 $\pm$ 0.04} & 2.64 $\pm$ 2.31$\ast$ \\
        \textbf{S11}    & 164.92 $\pm$ 14.84$\ast$ & \textbf{111.23 $\pm$ 43.51} & \textbf{229.00 $\pm$ 92.60} & 266.00 $\pm$ 67.13          & 0.04 $\pm$ 0.01          & 0.04 $\pm$ 0.01       & \textbf{0.07 $\pm$ 0.06} & 0.55 $\pm$ 0.91       \\
        \bottomrule
    \end{tabular}
    }
    \caption{Efficiency analysis of a swarm control agent with and without the presented context-awareness system.}
    \label{table:efficiency}
\end{table}

\subsection{Discussion}

We focus on synthesising results across the three lenses of mission effectiveness, control stability, and herding efficiency, drawing insight from the metrics discussed earlier in this section. Our first discussion considers the relationship between mission length and the total swarm distance moved. Considering each metric independently provides an understanding of the lenses we discussed previously. However, when compared directly, we observe where the shorter mission length is associated with further total distances moved, particularly where the control agent is augmented with the context-aware system. We found a strong negative correlation between mission length (all trials) and the swarm total distance moved for the control agent with context awareness ($r(9)=-0.8, p<0.01$) and no significant correlation for the control agent without context awareness ($r(9)=0.02, p>0.05$). Similar results for mission length (successful trials) and the total swarm distance moved. We found a strong negative correlation for the control agent with context-awareness ($r(9)=-0.77, p<0.01$) and no significant correlation for the control agent without context-awareness ($r(9)=-0.05, p>0.05$). The statistically significant correlation for the control agent with context awareness is unexpected as previous studies report that the likelihood of success decreases proportionally with the mission length~\cite{Strombom:2014}, suggesting that the longer the mission continues, the lower the success rate will be. Further investigation is required to understand the nature of the relationship between the swarm's total distance moved and the mission length.

One possible way to consider this outcome is to develop insight into this phenomenon when considering the different variable mission speeds. For the control agent with context awareness, there exists a strong negative correlation between mission length (all trials) and mission speed ($r(9)=-0.97, p<0.001$) and between mission length (successful trials) and mission speed ($r(9)=-0.95, p<0.001$), as well as a solid positive correlation between swarm total distance, moved and mission speed ($r(9)=0.78, p<0.01$). On the other hand, for the control agent without a context-aware system, these results hold in the case of all trials ($r(9)=-0.98, p<0.001$), are weak negative correlated for successful trials ($r(9)=-0.39, p>0.05$), although no significant correlation exists between the total swarm distance moved and mission speed ($r(9)=-0.06, p>0.05$). The strong negative correlation between mission speed and mission length metrics is expected for the control agent with context awareness. This intuitively can be interpreted that the faster an agent moves the swarm, the less time the mission will take.

We hypothesise that the difference between the control agent with and without context awareness is due, in part, to frequency behaviour modulation, set by $\sigma^{c_2}$ and $\sigma^{c_3}$. For the agent without context awareness, a \emph{stall distance} exists when $\beta$ becomes too close to any $\pi_i$~\cite{Perry2021:StallDist}. We introduce $\sigma^{c_3}$ to address the stall distance gap by maintaining the selected action modulation point invariant any $\pi_i$ response. As the control agent without context awareness does not have this capability and will stall when within a certain distance to a $\pi_i$, $\sigma^{c_3}$ could account for the incongruent observations between mission length and total swarm distance moved.

Our second point of discussion centres on the mean number of separated $\pi_i$ from the giant $\Pi$ cluster. Our initial expectation was that the control agent with context awareness would result in a statistically significantly lower (i.e. better) mean number of separated $\pi_i$ as a function of the number of decision changes the control agent makes. Recall that in the context-aware setting, $\sigma_2^{c_2}$ modulates the upper bound of this rate. Correlation analysis reveals a weak, non-significant correlation between these features for either the control agent with ($r(9)=0.30, p>0.05$) or without ($r(9)=0.30, p>0.05$) the context-awareness system. We postulate that the introduction of $\mathcal{L}$ in Equation~\ref{eqn:drive} may contribute to an increase of separated $\pi_i$ as we allow $\sigma_1$ behaviours to occur prior to the threshold $f(N)$ is achieved. What this means is that for the agent with the context-aware system, we change the decision boundary to allow a $\sigma_1$ action (drive) to occur \emph{prior} to the boundary condition $f(N)$ being met for all members of $\Pi$. While this does result in a greater number of separated $\pi$ members, the overall mission length (all trials and successful trials) is significantly reduced. This outcome is expected as separated $\pi$ may become isolated more easily for values of $\mathcal{L}\leq 0.75$ used in this study; investigation of systems with obstacles present is likely to provide further interesting results.

\section{Conclusions and Future Work}\label{sec:Conclusion}

Swarm control can be difficult due to the requirement to understand, control and anticipate agent responses. Typically, environmental complexities with static decision models are used to evaluate swarm control model performance. These explorations often maintain the standard baseline model to evaluate its performance through a particular lens, such as where this model breaks down. We assert that environmental influences that still impact the swarm agents continue to manifest as recognition and control problems for the swarm agent. By focusing on the agents of the swarm, their characteristics become the situations we identify. This enables us to determine the impact of the characteristic changes at individual or collective levels without specifying the properties of the surrounding environment.

In this paper, we have introduced a context-awareness system for a reactive shepherding control agent, demonstrating significant improvements in mission effectiveness. Our context-awareness system is an information marker-based approach to swarm control that focuses on the structuring and organisation of information to understand disparate contexts and situations of a swarm. The context-awareness system is a decision algorithm that focuses the attention of the swarm control agent on particular aspects of the swarm, reducing the search space of possible behaviours in response to the actions of the swarm. To enable this system, we conduct a systematic behaviour study that investigates the applicability of disparate control actions across distinct homogeneous and heterogeneous scenarios.

There are several avenues of future work to refine our context-awareness system, building on the architecture presented. The first is decision model selection. The implemented context-awareness decision model is a rule-based reasoning engine. Further research is required to select and evaluate alternative decision models suitable for robotic platform deployment, particularly in settings where sensor noise will perturb normal system operations. The second avenue of future work is to study the impact of incorrect context or situation identification. In this work, we use a closed system with a static environment, defined agent types and declared goal location. In real-world settings, this information may not always be available to an AI control agent, for instance, where a new type of agent in the swarm is present. In a shepherding setting, this could be the evaluation of heterogeneous flocks with agents not previously observed. A third avenue is to investigate the development of distributions of each metric for the context-aware control agent. This could provide the context-awareness system to assess the likelihood of success, given the context assessed and swarm state as measured by the metrics. The final avenue of future work is to consider adaptation and learning in the swarm and its effect on the control agent strategy. It is well established in biological settings that cognitive agents adapt and learn over time. However, these impacts remain open questions for swarm research.

\backmatter

\bmhead{Acknowledgements}

The authors wish to thank our research colleagues Kate J. Yaxley and Daniel P. Baxter for providing constructive feedback about the manuscript.

\section*{Declarations}

\bmhead{Funding}

The authors received no financial support for the research, authorship, and/or publication of this article.

\bmhead{Conflict of interest}

All authors declared that they have no conflicts for the research, authorship, and/or publication of this article.

\bmhead{Authors contributions}

A.H. conceptualised the study, designed and performed the experimentation (including model development, code implementation, and simulation output analysis), and prepared the manuscript. A.H., D.R. and H.A. supervised the study design, experimental conduct, and reviewed and commented on the manuscript at all stages.

\bibliography{bibliography}

\end{document}